\newcommand{\TitleFirstLine}{Incorporating Causal Graphical Prior Knowledge}
\newcommand{\TitleSecondLine}{into Predictive Modeling via Simple Data Augmentation}
\newcommand{\RunningTitle}{Incorporating Causal Graphical Prior Knowledge via Simple Data Augmentation}
\newcommand{\todo}[1]{}
\newcommand{\todoTwo}[1]{}
\documentclass{article}

\usepackage{arxiv}

\usepackage[utf8]{inputenc}
\usepackage[T1]{fontenc}
\usepackage{hyperref}
\usepackage{url}
\usepackage{booktabs}
\usepackage{amsfonts}
\usepackage{nicefrac}
\usepackage{microtype}
\usepackage{graphicx}
\usepackage{algorithm}
\usepackage[noend]{algpseudocode}\algrenewcommand\algorithmicdo{}

\usepackage[natbib=true,
style=ieee,
citestyle=numeric-comp,
maxcitenames=1, mincitenames=1,
maxbibnames=999,
url=false,isbn=false,doi=false]{biblatex}
\DefineBibliographyStrings{english}{andothers = {et al\adddot}
}
\AtEveryBibitem{\clearlist{language}
  \clearfield{series}
  \ifentrytype{book}{}{
    \clearfield{issn} \clearfield{doi} \clearlist{address}
      \clearfield{address}
      \clearfield{month}
      \clearname{editor}
      \clearfield{eprint}
      \clearlist{location}
      \clearfield{chapter}
  }
  \ifentrytype{online}{}{\clearfield{url}
  }
  \ifentrytype{inproceedings}{
\clearlist{publisher}
  }{}
}
\usepackage{amsmath,amsfonts,bm}
\usepackage{amssymb}
\usepackage{amsthm}
\newtheorem{theorem}{Theorem}

\newtheorem{proposition}{Proposition}
\newtheorem{lemma}{Lemma}
\newtheorem{definition}{Definition}
\newtheorem{assumption}{Assumption}
\newtheorem{remark}{Remark}
\newtheorem{fact}{Fact}

\usepackage{tikz}
\usetikzlibrary{shapes.geometric}
\usetikzlibrary {shapes.misc}
\usetikzlibrary{positioning}
\usepackage{lscape}
\usepackage{fontawesome}

\usepackage{subcaption}
\usepackage{enumerate}

\usepackage{mleftright}

\usepackage[toc,page]{appendix}

\author{
  Takeshi Teshima\\
  The University of Tokyo, RIKEN\\
  \texttt{teshima@ms.k.u-tokyo.ac.jp}\\
  \And
  Masashi Sugiyama\\
  RIKEN, The University of Tokyo\\
  \texttt{sugi@k.u-tokyo.ac.jp}
}

\hypersetup{
  pdftitle={Incorporating Causal Graphical Prior Knowledge into Predictive Modeling via Simple Data Augmentation},
  pdfsubject={},
  pdfauthor={Takeshi Teshima, Masashi Sugiyama},
  pdfkeywords={Causal graphical model, Data augmentation, Acyclic Directed Mixed Graph},
}

\date{}
 \newcommand{\DrawSubFig}[4]{\includegraphics[#4]{#1}\subcaption{#2}\label{#3}}
\def \figureRoot {.}
\newcommand{\indep}{{\!\perp\!\!\!\perp}}
\newcommand{\condIndep}[3]{#1\indep#2\ |\ #3}
\newcommand{\Equation}[1]{Eq.~\eqref{#1}}
\newcommand{\Figure}[1]{Fig.~#1}
\newcommand{\MSEBaseline}{\mathrm{MSE}_{\mathrm{base}}}
\newcommand{\MSEProposed}{\mathrm{MSE}_{\mathrm{prop}}}

\newcommand{\CompOrder}[1]{\mathcal{O}\left(#1\right)}

\newcommand{\ruleOfThumbBandwidth}{h^\mathrm{thumb}}

\DeclareRobustCommand{\coprod}{\mathop{\text{\fakecoprod}}}
\newcommand{\fakecoprod}{\sbox0{$\prod$}\smash{\raisebox{\dimexpr.9625\depth-\dp0}{\scalebox{1}[-1]{$\prod$}}}\vphantom{$\prod$}}
 \makeatletter
\let\save@mathaccent\mathaccent
\newcommand*\if@single[3]{\setbox0\hbox{${\mathaccent"0362{#1}}^H$}\setbox2\hbox{${\mathaccent"0362{\kern0pt#1}}^H$}\ifdim\ht0=\ht2 #3\else #2\fi
  }
\newcommand*\rel@kern[1]{\kern#1\dimexpr\macc@kerna}
\newcommand*\widebar[1]{\@ifnextchar^{{\wide@bar{#1}{0}}}{\wide@bar{#1}{1}}}
\newcommand*\wide@bar[2]{\if@single{#1}{\wide@bar@{#1}{#2}{1}}{\wide@bar@{#1}{#2}{2}}}
\newcommand*\wide@bar@[3]{\begingroup
  \def\mathaccent##1##2{\let\mathaccent\save@mathaccent
\if#32 \let\macc@nucleus\first@char \fi
\setbox\z@\hbox{$\macc@style{\macc@nucleus}_{}$}\setbox\tw@\hbox{$\macc@style{\macc@nucleus}{}_{}$}\dimen@\wd\tw@
    \advance\dimen@-\wd\z@
\divide\dimen@ 3
    \@tempdima\wd\tw@
    \advance\@tempdima-\scriptspace
\divide\@tempdima 10
    \advance\dimen@-\@tempdima
\ifdim\dimen@>\z@ \dimen@0pt\fi
\rel@kern{0.6}\kern-\dimen@
    \if#31
      \overline{\rel@kern{-0.6}\kern\dimen@\macc@nucleus\rel@kern{0.4}\kern\dimen@}\advance\dimen@0.4\dimexpr\macc@kerna
\let\final@kern#2\ifdim\dimen@<\z@ \let\final@kern1\fi
      \if\final@kern1 \kern-\dimen@\fi
    \else
      \overline{\rel@kern{-0.6}\kern\dimen@#1}\fi
  }\macc@depth\@ne
  \let\math@bgroup\@empty \let\math@egroup\macc@set@skewchar
  \mathsurround\z@ \frozen@everymath{\mathgroup\macc@group\relax}\macc@set@skewchar\relax
  \let\mathaccentV\macc@nested@a
\if#31
    \macc@nested@a\relax111{#1}\else
\def\gobble@till@marker##1\endmarker{}\futurelet\first@char\gobble@till@marker#1\endmarker
    \ifcat\noexpand\first@char A\else
    \def\first@char{}\fi
    \macc@nested@a\relax111{\first@char}\fi
    \endgroup
  }
  \makeatother 
\newcommand{\disjointUnion}{\coprod}

\usepackage{txfonts}
\newcommand{\spaceFont}[1]{\mathcal{#1}}

\providecommand{\annot}[2]{\underbrace{#1}_{\text{#2}}}
\newcommand{\bigtimes}{\times}
\newcommand{\argmin}{\mathop{\mathrm{arg~min}}\limits}
\newcommand{\Dirac}[1]{\delta_{#1}}
\newcommand{\absdet}[1]{\left|\det{#1}\right|}

\usepackage{dsfont}
\newcommand{\IndicatorSymbol}{\mathds{1}}
\newcommand{\Indicator}[1]{\mathop{\IndicatorSymbol}\!\left[#1\right]}

\newcommand{\dr}{\mathrm{d}}
\newcommand{\dt}{\dr{}t}
\newcommand{\du}{\dr{}u}
\newcommand{\D}{D}

\newcommand{\Z}{Z}
\newcommand{\Zs}{\mathbf{Z}}
\newcommand{\p}{p}
\newcommand{\pj}{\p_j}
\newcommand{\E}{\mathbb{E}}

\newcommand{\HypoCls}{\spaceFont{F}}
\newcommand{\f}{f}
\newcommand{\hf}{\hat{f}}
\newcommand{\jY}{{j^*}}
\newcommand{\X}{\mathbf{X}}
\newcommand{\Xsp}{\spaceFont{X}}
\newcommand{\runjX}{_{j \in [\D]\setminus\{\jY\}}}

\newcommand{\Ysp}{\spaceFont{Y}}

\newcommand{\Risk}{R}

\newcommand{\Data}{\mathcal{D}}
\newcommand{\hfemp}{\hat{f}_\mathrm{emp}}

\newcommand{\bi}{{\bm i}}

\newcommand{\lam}{\lambda}
\newcommand{\hRlam}{\tilde{R}_\lam}
\newcommand{\hRaug}{\hat{R}_{\mathrm{aug}}}

\newcommand{\hRemp}{\hat{R}_{\mathrm{emp}}}

\newcommand{\Reg}{\Omega}

\newcommand{\hpzmpj}{\hat{\p}_{j|\mpi{j}}}
\newcommand{\weightThreshold}{\theta}

\newcommand{\dZs}{\dr\Zs}
\newcommand{\dz}{\dr\z}

\newcommand{\Daug}{\mathcal{D}_\mathrm{aug}}
\newcommand{\runbi}{_{\bi \in [n]^\D}}
\newcommand{\Worig}{\mathcal{W}_\mathrm{orig}}
\newcommand{\Waug}{\mathcal{W}_\mathrm{aug}}

\newcommand{\setf}{\f \in \HypoCls}
\newcommand{\runf}{_{\setf}}

\newcommand{\fstar}{f^*}
\newcommand{\hRa}{\hRaug}
\newcommand{\supf}{\sup\runf}
\newcommand{\hpj}{\hat{p}_j}
\newcommand{\Zmpi}{\Zsijk{i}{\mpj}}
  
  \newcommand{\Zij}{\Z^{j}_i}
  \newcommand{\hw}{\hat{w}}
\newcommand{\wij}{\hw_i^j}
\newcommand{\wijj}{\hw_{i_j}^j}
\newcommand{\wbi}{\wiii{\bi}}
\newcommand{\wbij}{\wseqij{\bijm}{i_j}{j}}
\newcommand{\wbijm}{\wbijmk{1}{2}}

\newcommand{\wbijmk}[2]{\wseqij{\bijmk{#2}}{i_{j-#1}}{j-#1}}

\newcommand{\runfromj}[1]{_{#1=j+1}^\D{}}
\newcommand{\runtoj}[1]{_{#1=1}^{j-1}}
\newcommand{\hpk}[1]{\hat{p}_{#1}}
\newcommand{\runn}[1]{_{#1=1}^{n}}

\newcommand{\lf}{\ell_\f}
\newcommand{\lfj}{\ell_{\f,j}}
\newcommand{\lfjbi}{\ell_{\f,j}^\bijm}
\newcommand{\dzj}{\dr\zj}
\newcommand{\dzmp}{\dr\zmp}
\newcommand{\Zmpij}{\Zsijk{\bijm}{\mpj}}
\newcommand{\trivec}[3]{\begin{pmatrix} #1 \\ \vdots \\ #2 \\ #3 \end{pmatrix}}
\newcommand{\bivec}[2]{\trivec{#1}{#2}{}}

\newcommand{\hrj}{\hat{r}^j}
\newcommand{\hgj}{\hat{g}^j}

\newcommand{\gj}{g^j}
\newcommand{\rj}{r^j}

\newcommand{\pzmpjExt}{{\check\p}_{\mpi{j}}}
\newcommand{\pjmpjjointExt}{\check{\p}_{j,\mpi{j}}}

\newcommand{\setzmp}{\zmp \in \Zspmpj}
\newcommand{\runzmp}{_{\setzmp}}

\newcommand{\setbij}{\bijm \in [n]^{j-1}}
\newcommand{\runbij}{_{\setbij}}

\newcommand{\KernelShiftedClassHj}{\spaceFont{K}^j_{\mH}}
\newcommand{\Rademacher}[2]{\mathrm{Rad}_{#1,#2}}
\newcommand{\Radnp}{\Rademacher{n}{\p}}
\newcommand{\KRad}{\Radnp(\KernelShiftedClassHj)}
\newcommand{\ke}{k}
\newcommand{\runk}{_{\ke \in \KernelShiftedClassHj}}

\newcommand{\keBound}{B_\KAll}
\newcommand{\lossBound}{B_\ell}
\newcommand{\Erad}{\E_\rad}
\newcommand{\rad}{\sigma}
\newcommand{\Radmq}{\Rademacher{m}{q}}

\newcommand{\lFj}{{\bar{\spaceFont{L}}_\HypoCls^j}}
\newcommand{\lFjOne}{{\spaceFont{L}_\HypoCls^j}}
\newcommand{\setlfjOne}{\lfj' \in \lFjOne}
\newcommand{\runlfjOne}{_{\setlfjOne}}
\newcommand{\lFjOneKernel}{\lFjOne\otimes\KernelShiftedClassHj}
\newcommand{\lFjOneKRad}{\Radnp\left(\lFjOneKernel\right)}

\newcommand{\lF}{{\spaceFont{L}_\HypoCls}}

\newcommand{\convolution}[2]{#1 \mathbin{*} #2}
\newcommand{\convolutionFull}[3]{#1 \mathbin{\underset{[#3]}{*}} #2}

\newcommand{\CoveringNumber}[3]{\mathcal{N}_{\left(#2,#3\right)}(#1)}

\newcommand{\Deriv}[2]{\partial^{#1}\!{#2}}
\newcommand{\HolderName}{H\"{o}lder}

\newcommand{\HolderClass}[2]{\Sigma(#1, #2)}
\newcommand{\HolderClassFull}[3]{\Sigma(#1, #2, #3)}
\newcommand{\floor}[1]{{\lfloor #1 \rfloor}}
\newcommand{\zmpp}{{\zmp}'}

\newcommand{\setzj}{\zj \in \Zspj}
\newcommand{\runzj}{_{\setzj}}

\newcommand{\HolderBoundConst}[3]{\Phi(#1,#2,#3)}
\newcommand{\DiagMatrix}[1]{\mathrm{diag}(#1)}

\renewcommand{\Re}{\mathbb{R}}
\newcommand{\Na}{\mathbb{N}}
\newcommand{\Repos}{\Re_{> 0}}
\newcommand{\Renng}{\Re_{\geq 0}}
\newcommand{\Int}{\mathbb{Z}}
\newcommand{\Intnng}{\Int_{\geq 0}}
\newcommand{\MultiIndexSpace}[1]{\Intnng^{#1}}
\newcommand{\vect}[1]{{\bm{#1}}}
\newcommand{\mat}[1]{{\mathbf{#1}}}

\newcommand{\x}{\vect{x}}
\newcommand{\y}{\vect{y}}
\newcommand{\z}{\vect{z}}
\newcommand{\h}{\vect{h}}
\newcommand{\card}[1]{|#1|}

\newcommand{\ntom}[2]{[#1:#2]}

\newcommand{\nrm}[2]{\left\|#2\right\|_{#1}}
\newcommand{\opnrm}[1]{\nrm{\mathrm{op}}{#1}}
\newcommand{\supnrm}[1]{\nrm{\mathrm{\infty}}{#1}}
\newcommand{\twonrm}[1]{\nrm{}{#1}}

\newcommand{\Zsp}{\spaceFont{Z}}
\newcommand{\Zspk}[1]{\Zsp^{#1}}
\newcommand{\ZspkEntire}[1]{\widebar{\Zsp}^{#1}}
\newcommand{\hk}[1]{\h^{#1}}
\newcommand{\bikk}[2]{{\bi_{#1:#2}}}
\newcommand{\bij}{\bikk{1}{j}}
\newcommand{\bijmk}[1]{\bikk{1}{j-#1}}
\newcommand{\pk}[1]{\p_{#1}}
\newcommand{\zk}[1]{\z^{#1}}
\newcommand{\zmpk}[1]{\zk{\mpi{#1}}}
\newcommand{\Zijk}[2]{\Z^{#2}_{#1}}
\newcommand{\Zsijk}[2]{\Zs^{#2}_{#1}}
\newcommand{\Zi}[1]{\Zs_{#1}}
\newcommand{\Zk}[1]{\Zijk{}{#1}}
\newcommand{\Zsk}[1]{\Zs^{#1}}
\newcommand{\Zj}{\Zk{j}}
\newcommand{\Zmpj}{\Zsk{\mpj}}
\newcommand{\Zimpj}{\Zsijk{i}{\mpj}}
\newcommand{\pkmpk}[1]{\pk{#1|\mpi{#1}}}

\newcommand{\runjfrom}[1]{_{j=#1}^\D{}}
\newcommand{\runj}{\runjfrom{1}}
\newcommand{\runi}{_{i=1}^n}
\newcommand{\runij}{_{i_j=1}^n}

\newcommand{\Zspkk}[2]{{\Zspk{\ntom{#1}{#2}}}}

\newcommand{\zj}{\zk{j}}
\newcommand{\zmp}{\zmpk{j}}

\newcommand{\G}{\spaceFont{G}}
\newcommand{\hG}{\hat{\G}}
\newcommand{\Edges}{\spaceFont{E}}
\newcommand{\Biedges}{\spaceFont{B}}
\newcommand{\mpi}[1]{{\mathfrak{mp}(#1)}} \newcommand{\District}[1]{{\mathfrak{dis}(#1)}} \newcommand{\Parent}[1]{{\mathfrak{pa}(#1)}} \newcommand{\MarkovBlanket}[1]{{\mathfrak{mb}(#1)}}

\newcommand{\mpj}{\mpi{j}} \newcommand{\mpjsize}{{\card{\mpj}}}

\newcommand{\pjmpj}{\pkmpk{j}}
\newcommand{\pzmpj}{\pk{\mpi{j}}}
\newcommand{\pjmpjjoint}{\pk{j,\mpi{j}}}

\newcommand{\KernelSymbol}{K}
\newcommand{\Kj}{\KernelSymbol^j}
\newcommand{\tKj}{\tilde{\KernelSymbol}^j}

\newcommand{\Kkk}[2]{\KernelSymbol_{#2}^{#1}}
\newcommand{\mH}{\mat{H}}
\newcommand{\KAll}{\mat{K}}
\newcommand{\Hj}{\mH_j}
\newcommand{\KHj}{\KernelSymbol^j_{\mH}}

\newcommand{\invdetHj}{|\det\Hj|^{-1}}
\newcommand{\invHj}{\Hj^{-1}}

\newcommand{\Zspmpj}{\Zspk{\mpj}}
\newcommand{\Rempj}{\Re^\mpjsize}
\newcommand{\ZspAll}{\spaceFont{Z}}
\newcommand{\ReAll}{\Re^\D}

\newcommand{\Zspj}{\Zspk{j}}
\newcommand{\ZspjEntire}{\ZspkEntire{j}}
\newcommand{\ZspmpjEntire}{\ZspkEntire{\mpj}}

\newcommand{\wiij}[3]{\hw_{#2|#1}}
\newcommand{\wiiij}[2]{\wiii{#1}}
\newcommand{\wiii}[1]{\hw_{#1}}
\newcommand{\wseqij}{\wiij}
\newcommand{\bijm}{\bijmk{1}}
\newcommand{\Zbij}{\Z_\bijm}
\newcommand{\Zbijmpj}{\Zsijk{\bijm}{\mpj}}
\newcommand{\wbione}{\hw_{i_1}^1}

\newcommand{\IndNonzero}{I_{\neq 0}}

\newcommand{\ZspjBound}{B_j}

\newcommand{\epsmpj}{\epsilon_{\mpi{j}}}
\newcommand{\epsKHj}{\epsilon_{\Kj}(\Hj)}
\newcommand{\expandedDiffKHj}{\phi_{\Kj,\Hj}}
\newcommand{\ConstSumHj}{C_{\mH}}
\newcommand{\ConstSumIj}{C_{\p}}
\newcommand{\ConstMaxEpsK}{C_{\KAll}}
\newcommand{\RadHypoK}{R_{\HypoCls,\KAll}}
\newcommand{\RadK}{R_{\KAll}}

\newcommand{\BOne}{\frac{1}{\epsmpj}}
\newcommand{\BTwo}{\frac{\lossBound \keBound}{\epsKHj}}
\newcommand{\BThree}{\frac{\absdet{\Hj}}{\epsKHj}}
\newcommand{\BFour}{\frac{\lossBound\keBound}{\epsKHj}}
\newcommand{\BiasG}{\HolderBoundConst{\beta}{L}{\Kj} \opnrm{\Hj}^\beta + \expandedDiffKHj\pzmpjExtInt}
\newcommand{\BiasR}{\lossBound \ZspjBound \HolderBoundConst{\beta}{L}{\Kj} \opnrm{\Hj}^\beta + \lossBound\expandedDiffKHj\pjmpjjointExtInt}
\newcommand{\ProbR}{2\lFjOneKRad + \frac{\lossBound\keBound}{\absdet{\Hj}}\sqrt{\frac{\log(2/\delta)}{2n}}}
\newcommand{\ProbG}{2\KRad + \frac{\keBound}{\absdet{\Hj}}\sqrt{\frac{\log(2/\delta)}{2n}}}

\newcommand{\pzmpjExtInt}{\check{I}_{\mpi{j}}}
\newcommand{\pjmpjjointExtInt}{\check{I}_{j,\mpi{j}}}

\newcommand{\Asp}{\Zspmpj}
\newcommand{\setA}{\zmp \in \Asp}
\newcommand{\runA}{_{\setA}}

\newcommand{\nRounds}{K}
\newcommand{\nLeaves}{M}
\newcommand{\lTwoRegCoeff}{\rho}
\newcommand{\BinaryTree}[1]{\mathcal{T}_{#1}}
\newcommand{\GBRT}[2]{\HypoCls_{#1,#2}}
\newcommand{\Simplex}[1]{\Delta_{#1}}

\newcommand{\LipConst}[1]{L_{#1}}
\newcommand{\ZspRadiusBound}{R_\Zsp}
\newcommand{\EuclideanBall}[2]{\mathcal{B}^{#1}(#2)}
\newcommand{\Order}[1]{\mathcal{O}\left(#1\right)}
 
\newcommand{\tips}[1]{}
\def \Supplementary {Appendix}
\newcommand{\unitnum}[2]{#1\;#2}
\date{}
\title{\hspace*{\fill}\TitleFirstLine{}\hspace*{\fill}\newline{}\hspace*{\fill}\TitleSecondLine{}\hspace*{\fill}}
\begin{document}

\maketitle
\global\csname @topnum\endcsname 0

\begin{abstract}
Causal graphs (CGs) are compact representations of the knowledge of the data generating processes behind the data distributions.
When a CG is available, e.g., from the domain knowledge, we can infer the conditional independence (CI) relations that should hold in the data distribution.
However, it is not straightforward how to incorporate this knowledge into predictive modeling.
In this work, we propose a model-agnostic data augmentation method that allows us to exploit the prior knowledge of the CI encoded in a CG for supervised machine learning.
We theoretically justify the proposed method by providing an excess risk bound indicating that the proposed method suppresses overfitting by reducing the apparent complexity of the predictor hypothesis class.
Using real-world data with CGs provided by domain experts, we experimentally show that the proposed method is effective in improving the prediction accuracy, especially in the small-data regime.
\end{abstract}
\section{Introduction}
\label{sec:orga000ce4}
\label{paper:sec:intro}
\begin{figure}[t]
\begin{minipage}[t]{1.0\textwidth}\hspace*{\fill}
\begin{minipage}[c]{0.6\textwidth}
\begin{tabular}{cc}
\raisebox{-.5\height}{\includegraphics[keepaspectratio, width=0.65\linewidth]{\figureRoot/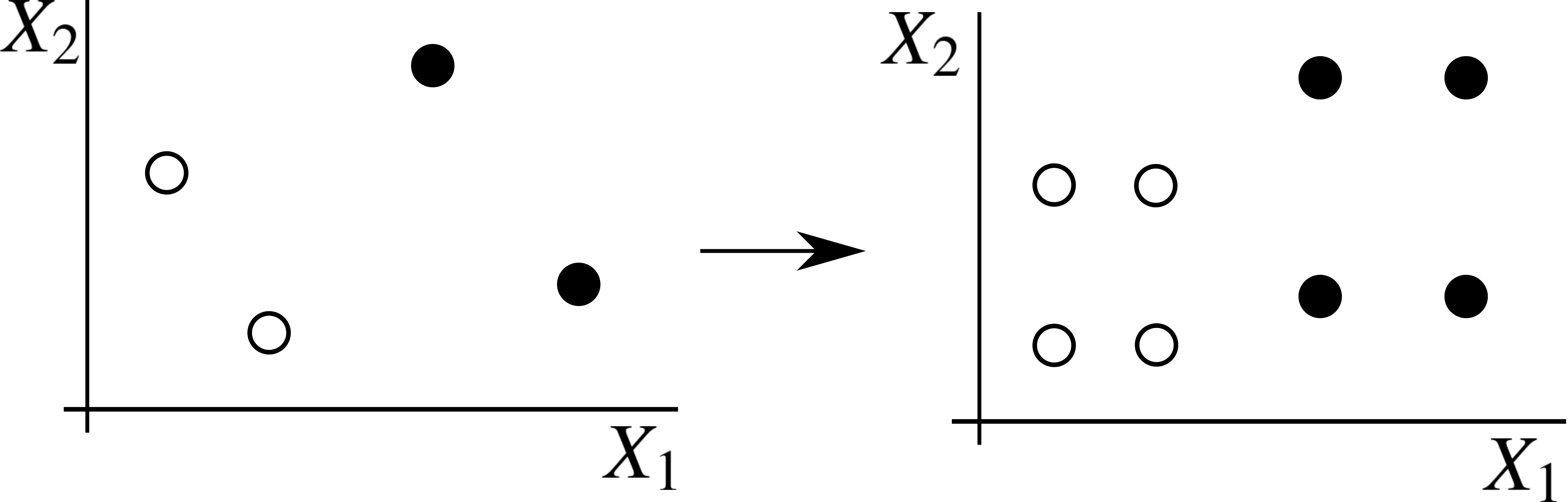}} & \(X_1 \gets Y \to X_2\)
\end{tabular}\hspace*{\fill}\\
\includegraphics[keepaspectratio, width=\linewidth]{\figureRoot/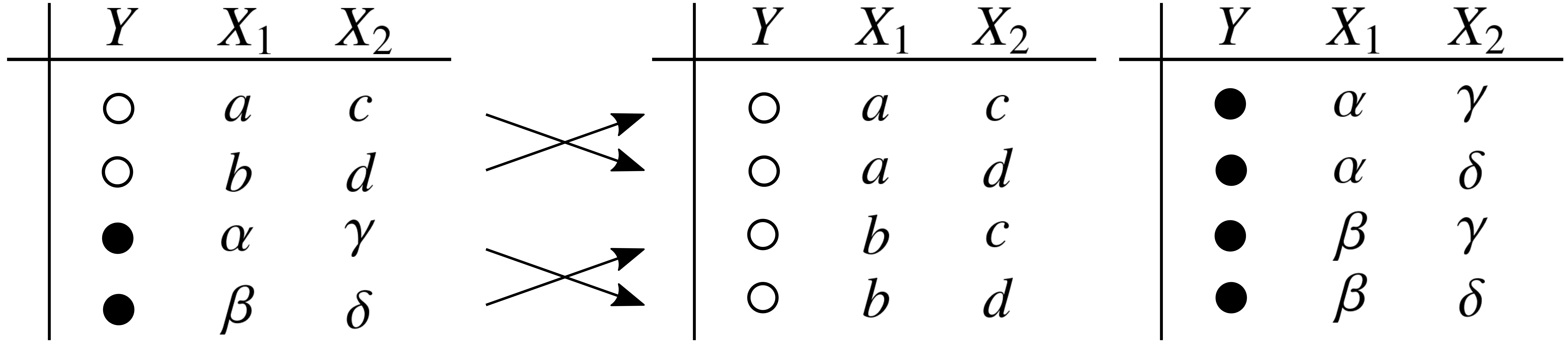}
\end{minipage}\hspace*{\fill}
\begin{minipage}[c]{0.3\textwidth}
\caption{
Visualization of the basic idea of the paper for the trivariate case \(X_1 \gets Y \to X_2\).
In this case, the CI \(\condIndep{X_1}{X_2}{Y}\) holds.
One way to use this knowledge via data augmentation is to group the data according to \(Y\) and then to shuffle \(X_1\) and \(X_2\) within each group.
Our method extends this idea to more general graphs.}
\label{fig:simple-example}
\end{minipage}\hspace*{\fill}
\end{minipage}
\end{figure}
\emph{Causal graphs} (CGs; \citealp{PearlCausality2009}) are compact representations of the knowledge of data generating processes.
Such a CG is sometimes provided by domain experts in some problem instances, e.g., in biology~\citep{SachsCausal2005} or sociology~\citep{ShimizuDirectLiNGAM2011}.
Otherwise, it may also be learned from data using the statistical causal discovery methods developed over the last decades~\citep{SpirtesCausation2000,PearlCausality2009,ChickeringOptimal2002,ShimizuLinear2006,PetersCausal2014,PetersElements2017}.
Once a CG is obtained, it can be used to infer the conditional independence (CI) relations that the data distribution should satisfy~\citep{PearlCausality2009}.

The CI relations encoded in the CG could be strong prior knowledge for
predictive tasks in machine learning, e.g., regression or classification, especially in the \emph{small-data} regime where data alone may be insufficient to witness the CI relations~\citep[Section~5.2.2]{SpirtesCausation2000}.
However, it is not trivial how the CI relations should be directly incorporated into general supervised learning methods.
In previous research, methods that leverage the causality for feature selection have been proposed (see, e.g., \citet{YuCausalitybased2020} for a review).
However, most of them are based on the notion of the \emph{Markov blanket} or the \emph{Markov boundary}~\citep{TsamardinosPrincipled2003}.
As a result, they only take into account partial information of all that is encoded in a CG,
since a CG often entails more constraints on the data distribution than the specifications of Markov blankets or a Markov boundary~\citep{RichardsonMarkov2003}.
Another approach to exploiting the prior knowledge of a CG is to build a \emph{Bayesian network} (BN) model according to the CG structure (e.g., \citealp{LucasBayesian2004}).
However, constructing the predictors by employing BNs as the framework entails a specific modeling choice,
e.g., it constructs a \emph{generative} model as opposed to a \emph{discriminative} model \citep[Chapter~24]{Shalev-ShwartzUnderstanding2014},
precluding the choice of some flexible and effective models such as tree-based predictors~\citep{FriedmanGreedy2001} and neural networks~\citep{GoodfellowDeep2016} that may be preferred in the application area of one's interest.

In this work, we propose a model-agnostic method to incorporate the CI relations implied by CGs directly into supervised learning via data augmentation.
To illustrate our idea, let us consider the following trivariate case.
\paragraph{Illustrative example: trivariate case (\Figure\ref{fig:simple-example}).}
\label{sec:orgedec376}
Suppose we want to predict a binary variable \(Y\) from \((X_1, X_2)\).
If the joint distribution follows the CG \(X_1 \gets Y \to X_2\), the CI \(\condIndep{X_1}{X_2}{Y}\) holds~\citep{PearlCausality2009}.
If we know this relation, a natural idea is to stratify the sample by \(Y\) and then to take all combinations of \(X_1\) and \(X_2\) within each stratum.

In this trivariate example, it is straightforward to derive such a plausible data augmentation procedure to incorporate the CI relations since the relation \(\condIndep{X_1}{X_2}{Y}\) involves all three variables.
On the other hand, deriving such a procedure for general graphs is not straightforward as they may encode a multitude of CI relations each of which may involve only a subset of all variables.
\paragraph{Our contributions.}
\label{sec:orgef89fe8}
(i)~We propose a method to augment data based on the prior knowledge expressed as CGs, assuming that an estimated CG is available.
(ii)~We theoretically justify the proposed method via an excess risk bound based on the Rademacher complexity~\citep{BartlettRademacher2002}.
The bound indicates that the proposed method suppresses overfitting at the cost of introducing additional complexity and bias into the problem.
(iii)~We empirically show that the proposed method yields consistent performance improvements especially in the small-data regime, through experiments using real-world data with CGs obtained from the domain knowledge.
\section{Problem Setup}
\label{sec:orgfd3d750}
\label{paper:sec:problem}
In this section, we describe the problem setup, the goal, and the main assumption exploited in our proposed method.
\paragraph{Basic notation.}
\label{sec:org2014375}
For the standard notation, namely \(\Re\), \(\Renng\), \(\Repos\), \(\Int\), \(\Na\), and \(\Indicator{\cdot}\), see Table~\ref{tbl:notation} in \Supplementary{} that also provides a summary of notation.
For \(N, M \in \Na\) with \(N \leq M\), define \([N:M] := \{N, N+1, \ldots, M\}\) and \([N] := [1:N]\).
For an \(N\)-dimensional vector \(\x = (x^1, \ldots, x^N)\) and \(S \subset [N]\), we let \(\x^S = (x^{s_1}, \ldots, x^{s_{|S|}})\) denote its sub-vector with indices in \(S = \{s_1, \ldots, s_{|S|}\}\) with \(s_1 < \cdots < s_{|S|}\). By abuse of notation, we write \(\x^j := \x^{\{j\}}\) for \(j \in [N]\).
To simplify the notation, we let \([0] = \emptyset\), \(\Re^0 := \{0\}\), \(\x^\emptyset = 0\), and \([N]^{0} = \{0\}\).
\paragraph{Problem setup and goal.}
\label{sec:org7a36e87}
Throughout the paper, we fix \(\D \in \mathbb{N}\), and let \(\Zsp = \bigtimes\runj \Zspk{j}\) where each \(\Zspk{j}\) is a subset of \(\ZspjEntire\) that is \(\Re\), \(\Int\), or a finite set.
Let \(\p\) be the joint probability density of \(\Zs := (\Z^1, \ldots, \Z^\D)\) taking values in \(\Zsp\).
One of the variables, e.g., \(\Zk{\jY} \ (\jY \in [\D])\), is the target variable that we want to predict.
Let \(\Xsp = \bigtimes\runjX\ZspkEntire{j}\) and \(\Ysp = \ZspkEntire{\jY}\).
Let \(\HypoCls \subset \Ysp^\Xsp\)
be a hypothesis class and \(\ell: \HypoCls \times \left(\bigtimes\runj \ZspkEntire{j}\right) \to \Re\) be a loss function.
We consider the supervised learning setting; that is, given the training data \(\Data = \{\Zi{i}\}_{i=1}^n\) that is an \mbox{independently} and identically distributed sample from \(\p\),
our goal is to find a predictor \(\hf \in \HypoCls\) with a small risk \(\Risk(\hf) = \E[\ell(\hf, \Zs)]\), where \(\E\) denotes the expectation with respect to \(\p\).

\paragraph{Assumption.}
\label{sec:orgf5aa901}
Let \(\G = ([\D], \Edges, \Biedges)\) be an \emph{acyclic directed mixed graph}\footnote{Here, \emph{mixed} indicates that the graph may contain bi-directed edges in addition to uni-directed ones.} (ADMG; \citealp{RichardsonMarkov2003,RichardsonNested2017}),
where \([\D]\) is the set of the vertices, \(\Edges\) is the uni-directed edges, and \(\Biedges\) is the bi-directed edges.
For the simplicity of exposition, in this paragraph, we temporarily assume that \([\D]\) is concordant with \emph{topological order} of \(\G\) without loss of generality.\footnote{That is, if \(1 \leq i < j \leq \D\), there is no directed path from \(j\) to \(i\).}
Our main assumption is that \(\p\) satisfies the \emph{topological ADMG factorization} property with respect to \(\G\)~\citep{BhattacharyaSemiparametric2020}, i.e.,
\begin{align}\label{eq:admg-factorization}
\p(\Zs) = \prod\runj \pjmpj(\Zj | \Zmpj),
\end{align}
where \(\mpi{j} \subset [j-1]\) denotes the \emph{Markov pillow} of \(j \in [\D]\) in \(\G\), and \(\pjmpj\) denotes the conditional density of \(\Zj\) given \(\Zmpj\).
The Markov pillow \(\mpi{j}\) is the collection of the following vertices: (1) those connected to \(j\) via bi-directed paths (including \(j\) itself), and (2) all parents of such vertices (see \Supplementary{}~\ref{sec:appendix:admg-terminology} or \citet{BhattacharyaSemiparametric2020} for the definition).
Markov pillow generalizes the notion of parents; if all edges are uni-directed, \(\mpi{j}\) matches the parents of \(j\),
and hence \Equation{eq:admg-factorization} is a generalization of the usual Markov factorization with respect to directed acyclic graphs (DAGs;~\citealp[p.16]{PearlCausality2009}) to ADMGs.
In the special case that the ADMG is \emph{uninformative}, i.e., when the graph is complete and all edges are bi-directed,
\Equation{eq:admg-factorization} reduces to the ordinary \emph{chain rule} of probability: \(\p(\Zs) = \prod\runj \p(\Zj|\Zsk{[j-1]})\), since \(\mpj = [j-1]\) in this case.
We assume that we are given an ADMG \(\hG = ([\D], \hat\Edges, \hat\Biedges)\) that is an estimator of \(\G\), and hereafter we assume that \([\D]\) is concordant with topological order of \(\hG\) without loss of generality.
\paragraph{Details on the assumption.}
\label{sec:org03f0c3e}
ADMGs with bi-directed edges appear in the case where unobserved confounders exist; they are used to represent \emph{semi-Markovian causal graphical models} (CGMs;~\citealp{TianGeneral2002}), which are CGMs allowing for the existence of hidden confounders.
The assumption of topological ADMG factorization is satisfied by such CGMs \citep{TianGeneral2002}.
We refer the readers to Section~2 of \citet{RichardsonNested2017} for an overview of ADMGs and their use in CGMs involving latent variables.
By accommodating not only DAGs (i.e., those without bi-directed edges) but also general ADMGs in the assumption, the applicability of the proposed method is extended to the case where there are unobserved confounders.
Note that the topological ADMG factorization, in general, captures only part of the equality constraints imposed by an ADMG on a semi-Markov model~\citep{BhattacharyaSemiparametric2020}.
Indeed, \citet{BhattacharyaSemiparametric2020} proposed a simple sufficient condition called the \emph{mb-shieldedness} (\emph{mb} stands for ``Markov blanket'') under which the topological ADMG factorization captures all the equality constraints.
Also note that a CG encodes more information/assumptions than the CI relations, namely, it encodes causal assumptions that describe how the data distribution should shift under an intervention \citep{PearlCausality2009}.
In this work, we only exploit the statistical assumptions, namely the CI relations, implied by a given CG.
Although our method does not directly exploit the causal interpretation of the DAGs/ADMGs, the causal modeling perspective can be useful in obtaining the DAGs/ADMGs from domain experts, i.e., one may be able to draw the DAGs/ADMGs by considering the (non-parametric) structural equations \citep{PearlCausality2009}.

\section{Proposed Method}
\label{sec:org306b237}
\label{paper:sec:proposed-method}
In this section, we explain the proposed data augmentation method to directly incorporate the prior knowledge of an ADMG into supervised learning.
The method generalizes the intuitive data augmentation method described in the trivariate DAG example in Section~\ref{paper:sec:intro},
making it applicable to general ADMGs whose encoded CI relations do not necessarily involve all variables.
The idea is to consider a \emph{nested conditional resampling}; instead of trying to generate all elements of the new data vector at once, we successively resample each variable from the \emph{conditional empirical distribution}~\citep{StuteConditional1986,HorvathAsymptotics1988} conditioning on its Markov pillow.
Then, our proposed method \emph{ADMG data augmentation} is obtained by considering all possible resampling paths simultaneously.
We later confirm that the proposed method indeed generalizes the previous procedure considered in the trivariate case of \Figure{\ref{fig:simple-example}}.
\paragraph{Derivation of the proposed method.}
\label{sec:org8609cb3}
Recall, given \Equation{eq:admg-factorization}, we can express the risk functional as
\begin{align*}
\Risk(f) &= \int_{\ZspAll} \ell(f, \Zs) \prod\runj \annot{\pjmpj(\Z^j | \Zmpj)}{(*)} \dZs.
\end{align*}
Then, to formulate the nested conditional resampling procedure, we select a kernel function \(\Kj : \ZspmpjEntire \to \Renng\) for each \(j \in [\D]\).\footnote{For notational simplicity, we define \(\Kj := 1\) where \(j\) is such that \(\mpi{j} = \emptyset\).}
Using this kernel function in the spirit of kernel-type function estimators~\citep{NadarayaEstimating1964,WatsonSmooth1964,EinmahlEmpirical2000},
we approximate each conditional density (\(\ast\)) as
\begin{align*}
\hpzmpj(\Zk{j} | \Zmpj) := \frac{\sum\runi\Dirac{\Z^j_i}(\Zk{j})\Kj(\Zmpj-\Zmpj_i)}{\sum_{k=1}^n \Kj(\Zmpj-\Zmpj_k)},
\end{align*}
where \(\Dirac{\z}\) denotes Dirac's delta function centered at \(\z\) (e.g.,~\citealp[Section~E.4.1]{ZorichMathematical2015}), and the right-hand side is defined to be zero when the denominator is zero.
The resulting approximation to the risk functional \(\Risk(f)\), denoted by \(\hRaug(\f)\), is
\begin{align*}
\hRaug(\f) := \int_{\ZspAll} \ell(f, \Zs) \prod\runj \hpzmpj(\Zk{j} | \Zmpj)\dZs.
\end{align*}
Here, the right-hand side can be interpreted as representing a nested conditional resampling procedure, in which we sequentially select \(i_1, \ldots, i_\D \in [n]\).
Indeed, since each \(\hpzmpj\) places its mass on \(\{\Z^j_i\}\runi\), the integration for \(\Zk{j}\) amounts to substituting \(\Zk{j} = \Z^j_{i_j}\) and summing over the choices \(i_j \in [n]\) with appropriate weights.
The weight placed on \(\Z^j_i\) by \(\hpzmpj\), namely \(\frac{\Kj(\Zmpj-\Zmpj_i)\IndNonzero}{\sum_{k=1}^n \Kj(\Zmpj-\Zmpj_k)}\), depends on \(\Zmpj\),
and it can be computed from \((\Z^1_{i_1}, \ldots, \Z^{j-1}_{i_{j-1}})\) which are already selected at the time we select \(\Z^j_{i_j}\) since \(\mpj \subset [j-1]\).
\paragraph{Proposed method.}
\label{sec:org33694b2}
By simultaneously considering all the possible resampling candidates, we reach at the \emph{instance-weighted data augmentation} procedure:
\begin{align}\label{eq:def-R-aug}
\hRaug(\f) = \sum_{\bi \in [n]^\D} \wbi \cdot \ell(f, \Zs_\bi),
\end{align}
where
\begin{align}\label{eq:def-wbi}
\wbi &= \prod\runj \frac{\Kj(\Zmpij-\Zimpj)}{\sum_{k=1}^n\Kj(\Zmpij-\Zsijk{k}{\mpj})}, \\
\Zs_\bi &= (\Zijk{i_1}{1}, \ldots, \Zijk{i_\D}{\D}),\quad
\Zsijk{\bijm}{} = (\Zijk{i_1}{1}, \ldots, \Zijk{i_{j-1}}{j-1}),\nonumber{}
\end{align}
for \(\bi = (i_1, \ldots, i_\D) \in [n]^\D\) and \(\bijm = (i_1, \ldots, i_{j-1})\),
and the right-hand side of \Equation{eq:def-wbi} is defined to be zero when the denominator is zero.
Here, we use the convention \(\Zsijk{\bikk{1}{0}}{\mpi{1}} := 0\) to be consistent with the notation.

Here, \Equation{eq:def-R-aug} represents a data-augmentation procedure in which new data points are created (see \Figure{\ref{fig:simple-example}}).
Each new data point \(\Zs_\bi\) is generated by the following procedure. First, \(D\) training data points are selected with replacement (specified by \(\bi = (i_1, \ldots, i_\D) \in [n]^D\)).
Then, \(\Zs_\bi\) is constructed by copying the \(j\)-th element \(\Zijk{i_j}{j}\) from \(\Zi{i_j}\) (\(j \in [\D]\)).
\Equation{eq:def-R-aug} performs this procedure for all combinations of the indices \(\bi \in [n]^D\).

In the proposed data augmentation method, which we call \emph{ADMG data augmentation}, we consider \(\Daug := \{\Zs_\bi\}\runbi\) to be a weighted training data whose weights are \(\Waug := \{\wbi\}\runbi\), and we perform supervised learning using \(\Daug\) and \(\Waug\), where any standard method that incorporates instance weights can be employed.
As a practical device, to account for the possibility that \(\hG\) is only an inaccurate approximation of \(\G\), we propose to use a convex combination of
the \emph{empirical risk estimator} \(\hRemp(f) := \frac{1}{n} \sum\runi \ell(\f, \Zs_i)\)
and the \emph{augmented empirical risk estimator} \(\hRaug(\f)\),
that is to use
\begin{align*}
\hf \in \argmin\runf \{(1 - \lam) \hRemp(\f) + \lam\hRaug(\f) + \Reg(\f)\}
\end{align*}
as the predictor, where \(\lambda \in [0, 1]\) is a hyper-parameter and \(\Reg\) is a regularization term for \(\f \in \HypoCls\).
In the experiments in Section~\ref{paper:sec:experiment}, we used a fixed parameter \(\lam = .5\) and observed that it performs reasonably for all data sets.

The ADMG data augmentation generalizes the idea described in the trivariate example \(X_1 \gets Y \to X_2\) in Section~\ref{paper:sec:intro}.
In fact, in the trivariate example of \Figure{\ref{fig:simple-example}}, \(\Waug\) places equal weights on the augmented data, essentially yielding the same augmented data set as that in \Figure{\ref{fig:simple-example}}.

\begin{figure}[t]
\begin{minipage}[c]{1.0\linewidth}\hspace*{\fill}
\begin{minipage}[c]{0.5\linewidth}
\hspace*{\fill}\includegraphics[keepaspectratio, width=1.0\linewidth]{\figureRoot/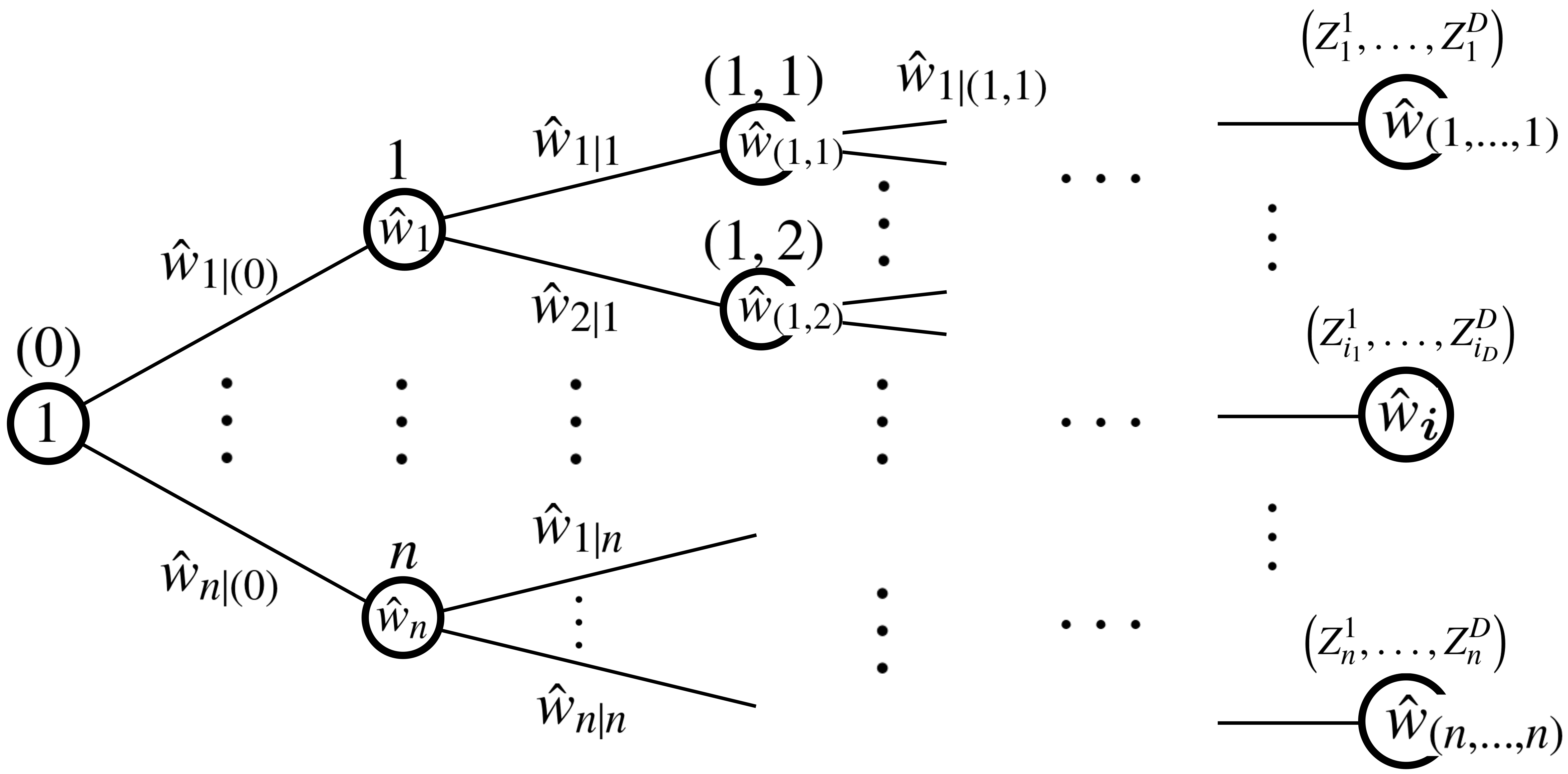}\hspace*{\fill}
\end{minipage}\hspace*{\fill}
\begin{minipage}[c]{0.4\linewidth}
\caption{Probability tree to compute the weights of the augmented instances.
At each depth \(j\), the index \(i_j\) is selected and the weight is updated as \(\wiiij{\bij}{j} = \wseqij{\bijm}{i_j}{j} \cdot \wiiij{\bijm}{j-1}\).
}
\label{fig:probability-tree}
\end{minipage}\hspace*{\fill}
\end{minipage}
\end{figure}

\paragraph{Practical implementation.}
\label{sec:orgf916aea}
To reduce the computation cost of calculating the weights \(\Waug\),
we exploit the recursive structure in \Equation{eq:def-wbi} that can be represented by a
probability tree~\citep{BraseUnderstanding2012}, where we sequentially select the values \(i_1, \ldots, i_\D \in [n]\)
(\Figure{\ref{fig:probability-tree}}).
To see this, recursively define
\begin{align*}
\wiii{\bikk{1}{0}} = 1,\quad \wiiij{\bij}{j} = \wseqij{\bijm}{i_j}{j} \cdot \wiiij{\bijm}{j-1} \ (j \in [\D], \bijm \in [n]^{j-1}),
\end{align*}
where
\begin{align*}
\wbij := \frac{\Kj(\Zbijmpj-\Zmpi)}{\sum\runi \Kj(\Zbijmpj-\Zmpi)},
\end{align*}
and the right-hand side is defined to be zero when the denominator is zero.
Then, we have \(\wbi = \wiii{\bikk{1}{\D}}\).

With this recursive structure in mind, we construct the probability tree as follows:
we index the root node by \(0\) and the nodes at depth \(j \in [\D]\) by \(\bij\) in a standard manner,
assign the weight \(\wbij\) to each edge \((\bijm, \bij)\),
and assign to each node \(\bij\) the product of the weights of the edges on the path from the root to \(\bij\).
Then, by recursively computing the weights of the nodes on this weighted tree, we can obtain \(\Waug\) (\Figure{\ref{fig:probability-tree}}).
Algorithm~\ref{paper:alg:proposed-method} summarizes the procedure of the proposed method.

To reduce the computation cost, we specify a threshold \(\weightThreshold \in (0, 1)\), and we prune the branches once the node weight becomes lower than \(\weightThreshold\) along the course of the recursive computation.
Since the edge weights satisfy \(\sum\runij\wbij \in \{0, 1\}\) and \(\wbij \geq 0\) for each \(\bijm\), the node weight \(\wiiij{\bij}{j}\) is monotonically decreasing in \(j\).
Therefore, the above pruning procedure only discards the nodes for which \(\wbi < \weightThreshold\).
The worst-case computational complexity of Algorithm~\ref{paper:alg:proposed-method} is \(\CompOrder{n^\D}\) (see \Supplementary{}~\ref{paper:sec:appendix:computational-complexity}), and it is important in future work to explore how to effectively reduce the computation complexity.
Apart from the pruning procedure, to reduce the computation time by taking advantage of the probability-tree structure, one may well consider employing heuristic top candidate search methods such as \emph{beam search}~\citep{BisianiBeam1987} or stochastic optimization methods such as \emph{stochastic gradient descent}~\citep[Section~5.9]{GoodfellowDeep2016}.

\begin{algorithm}[t]
\caption{
Proposed method: ADMG data augmentation
}\label{paper:alg:proposed-method}
\renewcommand{\algorithmicrequire}{\textbf{Input:}}
\renewcommand{\algorithmicensure}{\textbf{Output:}}
\begin{algorithmic}[1]
\Require Training data \(\Data\), ADMG \(\hG\), coefficient \(\lam \in [0, 1]\), regularization functional \(\Reg\), pruning threshold \(\weightThreshold \in [0, 1)\), hypothesis class \(\HypoCls\), kernel functions \(\{\Kj\}\runj\), loss function \(\ell\).
\Function{FillProbTree}{$\Data, \hG, \weightThreshold, \{\Kj\}\runj$}\Comment{see \Figure{\ref{fig:probability-tree}}}
\For{\(j \in [\D]\)} \Comment{for each variable $j$}

\For{\(\bijm \in [n]^{j-1}\)} \Comment{current node (depth $j$)}
\For{\(i_j \in [n]\)} \Comment{next node (depth $j+1$)}
\State \(\wiiij{\bijm}{j-1} \gets \wiiij{\bijm}{j-1} \Indicator{\wiiij{\bijm}{j-1} \geq \weightThreshold}\) \Comment{pruning}
\State \(\wiiij{\bij}{j} \gets \wseqij{\bijm}{i_j}{j} \cdot \wiiij{\bijm}{j-1}\)
\EndFor
\EndFor
\EndFor
\State \Return \(\Waug := \{\wbi\}\runbi\)
\EndFunction
\State Let $\Waug =$ \Call{FillProbTree}{$\Data, \hG, \weightThreshold, \{\Kj\}\runj$}.
\State Let $\hRaug(\f) := \sum_{\bi \in [n]^\D} \wbi \cdot \ell(\f, \Zs_\bi)$.
\State Let \(\hRlam(\f) := (1 - \lam) \hRemp(\f) + \lam\hRaug(\f) + \Reg(\f)\).
\Ensure \(\hf \in \argmin_{\f \in \HypoCls}\hRlam(\f)\): the predictor.
\end{algorithmic}\end{algorithm}
\section{Theoretical Justification}
\label{sec:orgd340a2c}
\label{paper:sec:theoretical-insights}
In this section, we provide a theoretical justification of the proposed method in the form of an excess risk bound, under the assumption that the CG is perfectly estimated.
The goal here is to elucidate how the proposed data augmentation procedure facilitates statistical learning from a theoretical perspective.
We focus on the case that \(\ZspjEntire = \Re\) for all \(j \in [\D]\).
Select \(\tKj\) and \(\h = (\hk{1}, \ldots, \hk{\D}) \in \Repos^\D\),
and define \(\Kj(u) := \frac{1}{|\det\Hj|}\tKj(\Hj^{-1} u)\), where \(\Hj := \DiagMatrix{\hk{\mpj}}\) is a diagonal matrix with elements \(\hk{\mpj}\).

For function classes, we quantify their complexities using the Rademacher complexity.
\begin{definition}[Rademacher complexity]
Let \(q\) denote a probability distribution on some measurable space \(\mathcal{X}\).
For a function class \(\mathcal{F} \subset \Re^\mathcal{X}\), define
\begin{align*}
\Radmq(\mathcal{F}) := \E_q\Erad \left[\sup_{f \in \mathcal{F}} \left|\frac{1}{m} \sum_{i=1}^m \rad_i f(X_i)\right|\right],
\end{align*}
where \(\{\rad_i\}_{i=1}^m\) are independent uniform \(\{\pm 1\}\)-valued random variables, and \(\{X_i\}_{i=1}^m \overset{\text{i.i.d.}}{\sim} q\).
\end{definition}
To state our result, let us define the set of marginalized functions and that of the shifted kernel functions as
\begin{align*}
&\lFjOne := \left\{\lfj(\zk{1}, \ldots, \zk{j-1}, \cdot): \f \in \HypoCls, (\zk{1}, \ldots, \zk{j-1})\in\Zspkk{1}{j-1}\right\},\\
&\left(\lfj : \bivec{\z^1}{\zj} \mapsto \int \ell(f, \z)\left(\prod\runfromj{k} \pkmpk{k}(\zk{k} | \zmpk{k})\right)\dz^{\ntom{j+1}{\D{}}}\right),\\
&\KernelShiftedClassHj := \left\{\Kj(\zmp-(\cdot)): \setzmp\right\},
\end{align*}
where the integration is over \(\Zspkk{j+1}{\D}\).
\begin{theorem}[Excess risk bound]
Let \(\hf \in \argmin\runf \{\hRaug(\f)\}\) and \(\fstar \in \argmin\runf \{\Risk(\f)\}\), assuming both exist.
Assume \(\hG = \G\) and also assume that \(\Zspj \subset \Re\) is compact.
Let \(\pzmpj\) and \(\pjmpjjoint\) denote the marginal density of \(\Zmpj\) and the joint density of \((\Zj, \Zmpj)\), respectively,
and assume \(\pzmpj\) and \(\pjmpjjoint(\zj, \cdot) \ (\zj \in \Zspk{j})\) have extensions to the entire \(\Rempj\) belonging to \(\HolderClass{\beta}{L}\), where \(\HolderClass{\beta}{L}\) denotes the \HolderName{} class of functions, \(\beta > 1\), and \(L > 0\).
Define
\begin{align*}
R_{\mH} &:= \sum\runj \left(\max_{j' \in \mpj}\hk{j'}\right)^\beta,\quad
R_{\KernelSymbol} := \sum\runj \absdet{\Hj} \KRad,\\
R_{\HypoCls,\KernelSymbol} &:= \sum\runj \absdet{\Hj} \lFjOneKRad.
\end{align*}
Under additional assumptions on the boundedness and smoothness of the kernels and the underlying densities (see Theorem~\ref{paper:thm:1:restatement} in \Supplementary{}~\ref{paper:sec:appendix:theory-1}),
there exist \(C_1, C_\p, C_2, C_3, C_4 > 0\) depending on the boundedness and the smoothness of \(\p, \ell, \{\tKj\}\runj\), and \(\mH\),
such that for any \(\delta \in (0, 1)\), we have with probability at least \(1 - \delta\),
\begin{align*}
\Risk(\hf) - \Risk(\fstar)
&\leq \annot{C_1 R_{\mH} + C_\p}{Kernel Bias}
+ \annot{C_2 R_{\KernelSymbol}}{Kernel Complexity}\\
&\qquad+ \annot{C_3 R_{\HypoCls,\KernelSymbol}}{Hypothesis Complexity}
+ \annot{C_4 \sqrt{\frac{\log(4\D{}/\delta)}{2n}}}{Uncertainty}.
\end{align*}
\label{paper:thm:1}
\end{theorem}
A proof is provided in \Supplementary{}~\ref{paper:sec:appendix:theory-1}.
Note that the existence of a smooth extension is satisfied by, e.g., a truncated version of a smooth density on \(\Rempj\).

\paragraph{Implications.}
Theorem~\ref{paper:thm:1} implies that the proposed method contributes to statistical learning by reducing the apparent complexity of the hypothesis class at the cost of introducing the additional complexity and bias arising from the kernel approximations.
In the interest of space, we provide a formal assessment of this complexity reduction effect in Proposition~\ref{prop:complexity-reduction} in \Supplementary{}~\ref{paper:sec:appendix:theory-2} under some additional Lipschitz-continuity assumptions.
In the derivation of Proposition~\ref{prop:complexity-reduction} indicating the complexity reduction effect, the fact that \(\lFjOne\) consists of univariate functions is critical.
In Section~\ref{paper:sec:experiment}, we empirically confirm that the complexity reduction effect is worth the newly introduced bias and complexity due to the kernel approximation in practice.

\paragraph{Scope of the analysis.}
It should be noted that the present theoretical guarantee only covers the case that the conditional independence assumptions implied by the CG are correct.
The robustness of the proposed method to the conditional independence assumptions is an important area of research in future work.
\section{Real-world Data Experiment}
\label{sec:org9c07fa5}
\label{paper:sec:experiment}
In this section, we report the results of the real-world data experiments to demonstrate the effectiveness of the proposed method in improving the prediction accuracy.

\subsection{Experiment Setup}
\label{sec:org5553b77}
\begin{figure*}[t]
\begin{minipage}[c]{1.0\linewidth}

\begin{minipage}[c]{1.0\linewidth}\hspace*{\fill}
\begin{minipage}[c]{0.48\linewidth}
\centering{}\includegraphics[keepaspectratio, height=0.2\textheight, width=\textwidth]{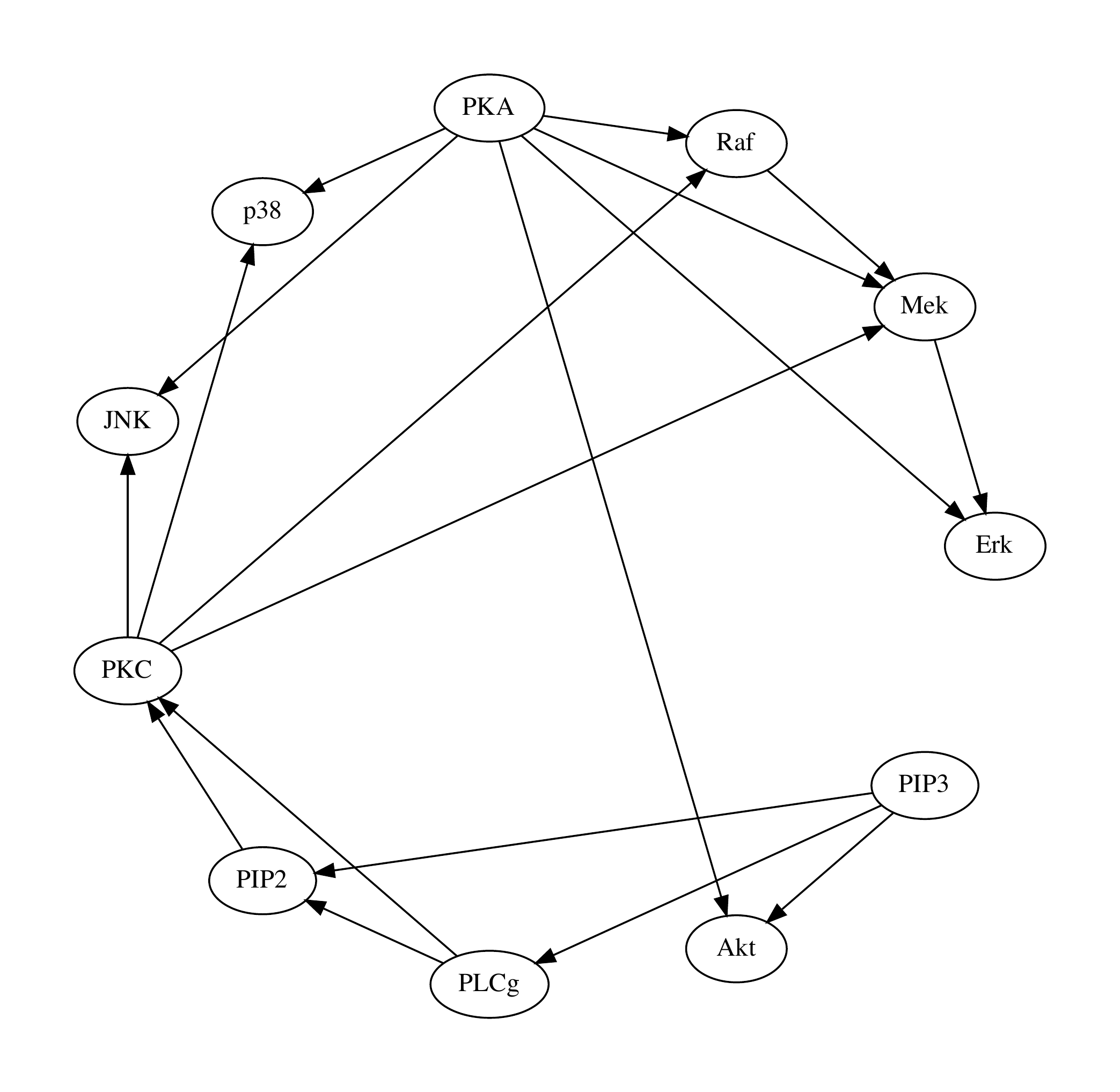}
\end{minipage}\hspace*{\fill}
\begin{minipage}[c]{0.48\linewidth}
\centering{}\includegraphics[keepaspectratio, height=0.2\textheight, width=0.8\textwidth]{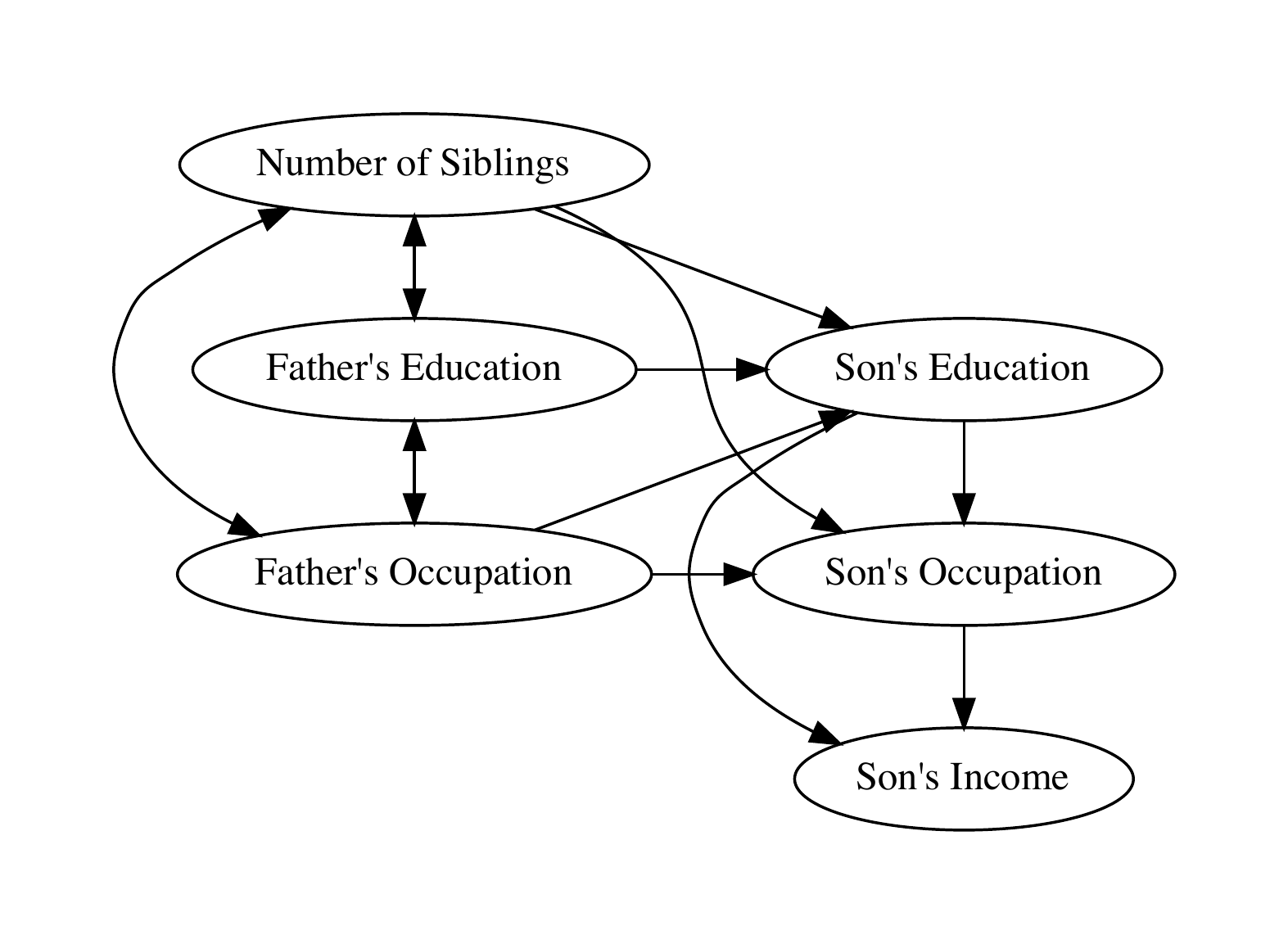}
\end{minipage}\hspace*{\fill}
\end{minipage}

\begin{minipage}[c]{1.0\linewidth}\hspace*{\fill}
\begin{minipage}[c]{0.45\linewidth}
\subcaption[sachs]{Reference graph for Sachs data.}\label{fig:sachs-graph}
\end{minipage}\hspace*{\fill}
\begin{minipage}[c]{0.45\linewidth}
\subcaption[gss]{Reference graph for GSS data.}\label{fig:gss-graph}
\end{minipage}\hspace*{\fill}
\end{minipage}

\end{minipage}

\begin{minipage}[c]{1.0\linewidth}
\caption{
Reference CGs for the data sets used in our experiments.
\subref{fig:sachs-graph} Consensus graph in \citet{SachsCausal2005}.
\subref{fig:gss-graph} Domain-knowledge graph based on the status attainment model \citep{DuncanSocioeconomic1972}.
}
\label{fig:reference-graph}
\end{minipage}
\end{figure*}

The goal of this experiment is to confirm that the proposed method contributes to the performance of the trained predictor, especially in the small-data regime.
To investigate the performance improvement, we make a comparison between the two cases: training with and without the proposed device, using the same hypothesis class and the same training algorithm.
To analyze the performance improvement in relation to the sample size, we vary the fraction of the data used for training the predictor and compare the performances of the proposed method and that of the baseline without a device.
For further details omitted here for the space limitation, please refer to \Supplementary{}~\ref{paper:sec:appendix:experiment-detail}.
\paragraph{Data sets.}
\label{sec:orga3d1652}
We employ 6 data sets for the experiment, namely
\emph{Sachs}~\citep{SachsCausal2005},
\emph{GSS}~\citep{ShimizuDirectLiNGAM2011},
\emph{Boston Housing}~\citep{HarrisonHedonic1978},
\emph{Auto MPG}~\citep{QuinlanCombining1993},
\emph{White Wine}~\citep{CortezModeling2009},
and \emph{Red Wine}~\citep{CortezModeling2009}.
Table~\ref{tbl:dataset} summarizes these data sets.
The \emph{Sachs} data and the \emph{GSS} data are accompanied by the ADMGs obtained from domain experts
(\Figure{\ref{fig:reference-graph}\subref{fig:gss-graph}} and \Figure{\ref{fig:reference-graph}\subref{fig:sachs-graph}}, respectively), and hence we use them in the experiment.
For the other data sets, we first perform \emph{DirectLiNGAM}~\citep{ShimizuDirectLiNGAM2011} on the entire data set to obtain the estimated CGs, simulating a situation that we have background knowledge from domain experts.
Since DirectLiNGAM produces DAGs, the CGs used in this experiment are DAGs except for the case of \emph{GSS} data set which is accompanied by an ADMG produced by domain experts
(\Figure{\ref{fig:reference-graph}\subref{fig:gss-graph}}).
\paragraph{Predictor model class.}
\label{sec:orgdbc4ed8}
We employ the gradient boosted regression trees~\citep{FriedmanGreedy2001,ChenXGBoost2016} as the predictor model class.
The hypothesis class consists of the convex combinations of binary regression trees with at most \(\nLeaves\) leaves:
\begin{align*}
\GBRT{\nLeaves}{\nRounds} := \left\{\sum_{k=1}^{\nRounds} \alpha^k w_k^{h_k(\cdot)} : \alpha \in \Simplex{\nRounds}, T_k \in [\nLeaves], w_k \in \Re^{T_k}, h_k \in \BinaryTree{T_k}\right\},
\end{align*}
where \(\nLeaves, \nRounds \in \Na\), \(\BinaryTree{T}\) represents the set of binary tree structures mapping \(\Xsp\) to \([T]\),
and \(\Simplex{\nRounds}\) is the \((\nRounds-1)\)-dimensional probability simplex.
The loss function is the squared error \(\ell(\f, \Zs) = (Y - \f(\X))^2\) where \(Y = \Zk{\jY}\) and \(\X = \Zsk{[\D]\setminus\{\jY\}}\),
and the regularization function is \(\Omega(\f) = \sum_{k=1}^{\nRounds} \frac{\lTwoRegCoeff}{2}\twonrm{w_k}^2 (\lTwoRegCoeff > 0)\).
We fix \(\nLeaves = 64\) and search the number of boosting rounds \(\nRounds\) in \(\{10, 50, 250, 1250\}\) and the \(\ell_2\)-regularization coefficient \(\lTwoRegCoeff\) in \(\{1, 10, 100, 1000\}\).
The hyper-parameters are selected by the grid-search based on 3-fold weighted cross-validation.
Note that, for the proposed method, we perform cross-validation on the union of the original training data and the augmented data with the weights adjusted by \(\lam\), namely \(\Data \disjointUnion \Daug\) with weights \((1 - \lam) \Worig \disjointUnion \lam \Waug\) where \(\Worig = (\frac{1}{n}, \ldots, \frac{1}{n})\).
\paragraph{Configurations of the proposed method.}
\label{sec:org739c605}
We select \(\h = (\hk{1}, \ldots, \hk{\D}) \in \Repos^\D\) and use the product kernel \(\Kj(\x-\y) := \prod_{j' \in \mpj}\frac{1}{\hk{j'}}\Kkk{j}{j'}\left(\frac{\x^{j'}-\y^{j'}}{\hk{j'}}\right)\) for the proposed method.
For each \(j' \in \mpj\), if the variable is continuous (i.e., \(\ZspkEntire{j'} = \Re\)), we use the Gaussian kernel \(\Kkk{j}{j'}(x - y) := (2\pi)^{-1/2}\exp\left(-\frac{(x - y)^2}{2}\right)\).
Otherwise, i.e., if the variable is discrete, we use the identity kernel \(\Kkk{j}{j'}(x-y) := \Indicator{x = y}\) and \(\hk{j'} = 1\).
For the Gaussian kernels, we select the \emph{kernel bandwidth} \(\hk{j'}\) based on \emph{Silverman's rule-of-thumb}~\citep[pp.45--47]{SilvermanDensity1986}.
In the experiment, we fix \(\lam = .5\) throughout all runs and find that it yields reasonable performances in all data sets.
\paragraph{Compared methods.}
\label{sec:orgd3ee7f4}
We compare the performances of the proposed method
and the naive baseline method without a device:
\begin{align*}
\hf \in \argmin\runf \{\hRemp(\f) + \Reg(\f)\}.
\end{align*}
In Section~\ref{paper:sec:experiment:results} where we report the results,
the two methods are referred to as \emph{Proposed} and \emph{Baseline}, respectively.
\paragraph{Evaluation procedure.}
\label{sec:orga6318ec}
The prediction accuracy is measured by the mean squared error (MSE).
For each data set, we randomly subsample a fraction of the data as the training set and use the rest as the testing set.
The fraction of the training set is varied in \(\{.1, .15, \ldots, .85\}\).
For each training set fraction, random train-test splits are performed \(20\)~times.
Subsequently, for each split, \emph{Proposed} and \emph{Baseline} are trained on the training set, and then evaluated on the testing set.
We report the average performances as well as the standard errors over the \(20\) runs for each training set fraction.
\begin{table}[t]
\caption{
Summary of Data Sets
(\emph{NAME}: name of the data set,
\emph{\#VAR}: number of variables in the data set,
\emph{\#OBS}: number of observations,
\emph{GRAPH}: CG used for the proposed method,
\emph{Consensus}: consensus network
(\Figure{\ref{fig:reference-graph}\subref{fig:gss-graph}}),
\emph{Domain}: domain knowledge of the status attainment model
(\Figure{\ref{fig:reference-graph}\subref{fig:sachs-graph}}),
\emph{LiNGAM}: CG is estimated by performing DirectLiNGAM on the entire data set).
}
\label{tbl:dataset}
\begin{center}
\begin{tabular}{lrrl}
\multicolumn{1}{c}{\bf NAME}  &\multicolumn{1}{c}{\bf \#VAR} &\multicolumn{1}{c}{\bf \#OBS} &\multicolumn{1}{c}{\bf GRAPH}\\
\hline \\
\emph{Sachs} & 11 & 853 & Consensus \\
\emph{GSS} & 6 & 1380 & Domain \\
\emph{Boston Housing} & 14 & 506 & LiNGAM \\
\emph{Auto MPG} & 7 & 392 & LiNGAM \\
\emph{White Wine} & 12 & 4898 & LiNGAM \\
\emph{Red Wine} & 12 & 1599 & LiNGAM \\
\end{tabular}
\end{center}
\end{table}
\subsection{Results}
\label{sec:orge697090}
\label{paper:sec:experiment:results}
\begin{figure*}[!ht]
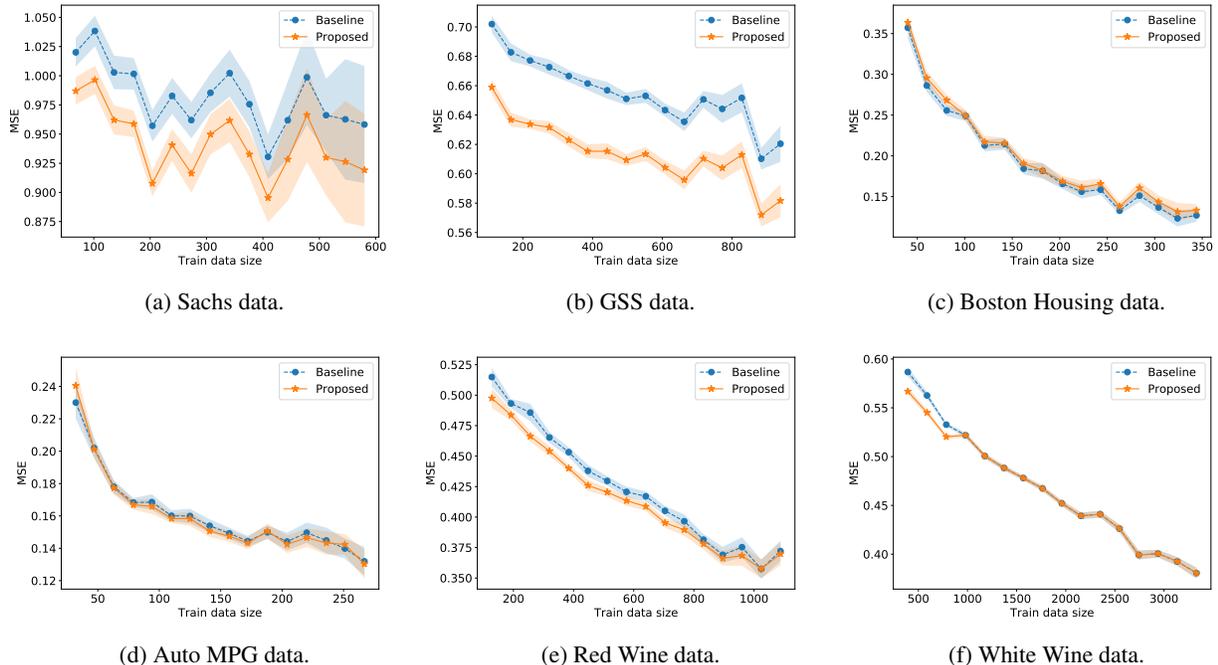

\begin{minipage}[c]{1.0\linewidth}
\begin{minipage}[c]{0.33\linewidth}
\centering{}\DrawSubFig{\figureRoot/sachs}{Sachs data.}{fig:sachs}{keepaspectratio, height=\textheight, width=\textwidth}
\end{minipage}\hfill
\begin{minipage}[c]{0.33\linewidth}
\centering{}\DrawSubFig{\figureRoot/gss}{GSS data.}{fig:gss}{keepaspectratio, height=\textheight, width=\textwidth}
\end{minipage}\hfill
\begin{minipage}[c]{0.33\linewidth}
\centering{}\DrawSubFig{\figureRoot/boston_housing}{Boston Housing data.}{fig:boston_housing}{keepaspectratio, height=\textheight, width=\textwidth}
\end{minipage}\hfill
\end{minipage}

\begin{minipage}[c]{1.0\linewidth}
\begin{minipage}[c]{0.33\linewidth}
\centering{}\DrawSubFig{\figureRoot/auto_mpg}{Auto MPG data.}{fig:auto_mpg}{keepaspectratio, height=\textheight, width=\textwidth}
\end{minipage}\hfill
\begin{minipage}[c]{0.33\linewidth}
\centering{}\DrawSubFig{\figureRoot/red_wine}{Red Wine data.}{fig:red_wine}{keepaspectratio, height=\textheight, width=\textwidth}
\end{minipage}\hfill
\begin{minipage}[c]{0.33\linewidth}
\centering{}\DrawSubFig{\figureRoot/white_wine}{White Wine data.}{fig:white_wine}{keepaspectratio, height=\textheight, width=\textwidth}
\end{minipage}\hfill
\end{minipage}

\begin{minipage}[c]{1.0\linewidth}
\caption{
Illustration of the experimental results.
In all figures, the horizontal axis is the varied size of the training data before augmentation,
and the vertical axis is the performance metric (MSE; the lower the better).
The markers and the lines indicate the average over the 20 independent runs, and the shades are drawn for the width of the standard errors both above and below the lines.
The proposed method shows a consistent improvement over the naive baseline based on the empirical risk minimization with the same hypothesis class, particularly in the small-data regime.
}
\label{fig:result-1}
\end{minipage}
\end{figure*}

\Figure{\ref{fig:result-1}} shows the experimental result.
We observe a consistent performance improvement in most of the data sets.
For the data sets for which the domain knowledge CG is provided (i.e., \emph{Sachs} and \emph{GSS}), we can see clear relative improvement ranging from 3\% to 7\% on average, especially in the small-data regime where approximately 10--40\% is the training set fraction.
In the other data sets without the background knowledge, relatively little improvement is observed except in the small-data regions of \emph{Red Wine} and \emph{White Wine}, where up to 4\% relative improvement on average is observed.
The lack of relative improvement in the majority of these cases emphasizes the importance of having accurate domain knowledge in the proposed approach, and it motivates the development of effective causal discovery methods.
In the \emph{White Wine} data, the proposed method coincides with the baseline in the larger-data region as the augmentation did not effectively take place due to the adaptive bandwidth that is narrowed according to the sample size.
For supplementary figures visualizing the average relative improvements, see \Supplementary{}~\ref{paper:sec:appendix:experiment-relative-improvement}.
\section{Related Work and Discussion}
\label{sec:org0990bcb}
\label{paper:sec:related-work}
In this section, we explain the context of the paper in relation to existing work.
\subsection{CGMs and Predictive Modeling}
\label{sec:orga4ac7f2}
\paragraph{Variable selection in a single-distribution setting.}
\label{sec:orge1b40dc}
The background knowledge encoded in a CG can be used for variable selection by identifying a \emph{Markov boundary} of the target variable.
Here, \(\MarkovBlanket{j} \subset [\D]\) is called a \emph{Markov blanket} of \(j\) if \(\Z^j\) is conditionally independent of all the other variables given \(\Zsk{\MarkovBlanket{j}}\).
If, moreover, \(\MarkovBlanket{j}\) is minimal, i.e., if none of its proper subsets are Markov blankets, it is called a \emph{Markov boundary} (MB).
Under certain assumptions, the MB of a target variable is known to be the minimal set of variables with optimal predictive performance~\citep{TsamardinosPrincipled2003}.
For a recent comprehensive review on MB estimation, see \citet{YuCausalitybased2020}.
The present paper is orthogonal to this line of work.
In fact, the CGs can encode more information than a specification of the Markov boundary of the predicted variable; for example, consider the CG \(X_1 \gets Y \to X_2\) where \(Y\) is the target variable and \((X_1, X_2)\) are the predictors.
In this case, the Markov boundary of \(Y\) is \((X_1, X_2)\), and hence the variable selection does not reduce the number of the predictors.
On the other hand, the proposed method still leverages the factorization structure of the data distribution entailing the CG.
In practice, the two approaches can be combined straightforwardly.
In our experiments, we do not perform variable selection using the data regarding the possibility that the obtained CGs are inaccurate.
\paragraph{Variable selection in distribution-shift setting.}
\label{sec:orgc1ee810}
Another line of research is concerned with making predictions under distribution shift and leverage feature selection based on causal background knowledge or causal discovery.
\citet{MagliacaneDomain2018} considered the case that a distribution shift is due to intervention in some variables, and they proposed a method to perform domain adaptation by identifying a set of variables that is likely to perform well regardless of the intervention.
\citet{Rojas-CarullaInvariant2018} assume that if the conditional distribution of the predicted variable given some subset of features is invariant across different distributions, then this conditional distribution is the same in the \emph{target distribution} for which one wants to make good predictions,
and leveraged it to find the set of variables for which the relation to the target variable does not change.
The present paper is complementary to this line of work since our goal is to make good predictions in a single fixed distribution.
\paragraph{Regularization and model selection.}
\label{sec:orgdb94853}
\citet{KyonoImproving2019} proposed a model selection criterion that can reflect the structure of a CG.
The goal of \citet{KyonoImproving2019} is \emph{domain generalization} and \emph{out-of-distribution prediction}, i.e., making good predictions under a distribution shift without access to any samples from the target distribution or making good predictions for the data that is outside the support of the training data distribution.
To achieve it, given a DAG as prior knowledge, \citet{KyonoImproving2019} first modify it so that the edges coming out of the target variable are removed.
Then, to score the predictor model candidates, it generates a data set whose predicted variables are replaced by the predictions of the model and computes the \emph{Bayes Information Criterion} (BIC) that evaluates the fitness of the modified DAG structure to the generated data set.
Another approach for using the background knowledge of a CG is the \emph{CASTLE regularization}~\citep{KyonoCASTLE2020}.
CASTLE regularization regularizes a neural network while performing the CG discovery as an auxiliary task.
The method imposes a reconstruction loss using the internal layers of the predictor implemented by neural networks under a DAG constraint.
The present paper is orthogonal to these researches and can be straightforwardly combined in practice.
Also note that our method has a theoretical justification while \citet{KyonoImproving2019} provided no theoretical justifications.

\paragraph{Inference under specific CGs.}
\label{sec:orgffe806f}
Under some specific problem settings with known specific underlying CGs, methods to take advantage of the prior knowledge have been developed.
For example, in the instance weight estimation for episodic reinforcement learning, methods to perform \emph{state simplification} based on the CGs
have been proposed~\citep[Section~8.2]{BottouCounterfactual2013,PetersElements2017}.
\citet{ScholkopfRemoving2015} considered removing systematic errors using \emph{half-sibling regression} inspired by the CG of the observation mechanism found in the \emph{exoplanet search}.
\citet{PitisCounterfactual2020} proposed a method to enhance the sample efficiency in reinforcement learning (RL) by a procedure to exchange the realizations of the variables within the (conditionally) disconnected components in the CG of the \emph{Markov decision process} of specific RL instances.
This line of work and the present work are complementary in that our approach is widely applicable to general ADMGs whereas these analyses have the potential to exploit the characteristics of the specific problem setups.
\paragraph{Causal bootstrapping.}
\label{sec:orgd7c1472}
Recently, \citet{LittleCausal2020} proposed \emph{causal bootstrapping}, a weighted bootstrap-type algorithm that is relevant to our method.
While, methodologically, both the present paper and \citet{LittleCausal2020} can be seen to be based on kernel-type function estimators~\citep{StuteConditional1986,HorvathAsymptotics1988,EinmahlEmpirical2000} and CGs~\citep{PearlCausality2009},
the two works are complementary in that the problem setups differ.
Causal bootstrapping of \citet{LittleCausal2020} aims at mitigating the performance degradation due to a distribution shift arising from an intervention, and it uses kernel-type function estimators to simulate sampling from an interventional distribution.
On the other hand, we investigate the performance improvement yielded from using the background knowledge of a CG in a scenario without a distribution shift.
\paragraph{Constructing probabilistic graphical models.}
\label{sec:orgf2ec9de}
\citet{EvansMarkovian2014} provided a smooth parametrization of the set of distributions that are \emph{Markov with respect to} an ADMG \(\G\) in the binary case: \(\ZspkEntire{j} = \{0, 1\} \ (j \in [\D])\).
Complementarily, for the case of \(\ZspkEntire{j} = \Re \ (j \in [\D])\), \citet{SilvaMixed2011} proposed the construction of flexible probability models that are Markov with respect to a given ADMG.
Similarly, in the case that the ADMG has no bi-directed edges, constructing a Bayesian network by specifying the conditional distributions appearing in the Markov factorization (\Equation{eq:admg-factorization}) is one natural way to exploit this prior knowledge~\citep{LucasBayesian2004}.
This approach has the limitation that it inevitably restricts the modeling choice, e.g., the constructed predictor is a generative model as opposed to a discriminative model \citep[Chapter~24]{Shalev-ShwartzUnderstanding2014}, whereas our approach has the virtue of being model-agnostic.
\subsection{Causal Discovery and Transfer Learning}
\label{sec:org8f73c70}
Our method provides a channel through which an estimated CG can be used for enhancing the predictive modeling.
In this sense, the proposed method can serve as a transfer learning method under a \emph{transfer assumption} of \emph{common CG}, i.e.,
an assumption that one is given many samples from another distribution sharing the same CG with the distribution for which we want to make the predictions.
Under such an assumption, one may first estimate the ADMG using causal discovery methods to estimate the \emph{Markov equivalence class} of ADMGs expressed as a \emph{partial ancestral graph} (PAG)~\citep{ZhangCompleteness2008}, e.g., the \emph{fast causal inference} (FCI) algorithm~\citep{SpirtesCausal1995,ZhangCompleteness2008},
enumerate the ADMGs in the equivalence class (e.g., by the \emph{Pag2admg} algorithm; \citealp{SubramaniPag2admg2018}),
select a plausible candidate ADMG that is concordant with the domain knowledge, and apply the proposed method.
Such an assumption of a common causal mechanism has been exploited in recent work of causal discovery~\citep{XuPoolingLiNGAM2014,GhassamiLearning2017,MontiCausal2019} and transfer learning~\citep{PearlTransportability2011,MagliacaneDomain2018,TeshimaFewshot2020}, and it is based on a common belief that a causal mechanism remains invariant unless explicitly intervened in~\citep{HunermundCausal2019}.
\subsection{CGMs and Efficient Estimation}
\label{sec:org0c83de2}
Our method could be also seen as a method to perform sample-efficient inference given a CG.
In the existing work, the knowledge of a CG has been used for deriving efficient estimators for \emph{identifiable causal estimands} \citep{PearlCausality2009} such as the \emph{interventional distributions} \citep{JungEstimating2021,JungEstimating2021a} or the \emph{average causal effect} \citep{BhattacharyaSemiparametric2020}.
For instance, \citet{JungEstimating2021} and \citet{JungEstimating2021a} derived expressions of efficient estimators of the identifiable interventional distributions given an ADMG and a PAG, respectively, by leveraging the knowledge of the CG in the \emph{double/debiased machine learning} \citep{ChernozhukovDouble2018} framework.
Another line of research provided graphical criteria for selecting the \emph{efficient adjustment sets}, the set of covariates to be adjusted for producing a valid estimator of a causal effect with the minimal asymptotic variance \citep{HenckelGraphical2020,RotnitzkyEfficient2020,WitteEfficient2020,SmuclerEfficient2021}.
Our goal differs from the goals of these lines of research; we are interested in improving the sample efficiency of training the predictor whereas they aimed to improve the sample efficiency of causal inference.
Nevertheless, it is an interesting direction of future research to elucidate whether the proposed method is optimally efficient in estimating the risk functional given the CG.
\section{Conclusion}
\label{sec:org38dc411}
\label{paper:sec:conclusion}
In this paper, we proposed a general method for exploiting the causal prior knowledge in predictive modeling.
We theoretically provided an excess risk bound indicating that the proposed method has a complexity reduction effect that mitigates overfitting while it introduces additional complexity and bias arising from the kernel approximations.
Through the experiments using real-world data, we demonstrated that the proposed method consistently improves the predictive performance especially in the small-data regime,
which implies that the complexity reduction effect is worth the newly introduced bias and complexity in practice.
Important areas in future work include
incorporating the equality constraints imposed by an ADMG but not captured by the topological ADMG factorization
and handling more relaxed assumptions such as those expressed as PAGs.
\section*{Acknowledgments}
\label{sec:org4753097}
The authors are grateful to Prof.~Shohei Shimizu for providing them with the preprocessed GSS data set used in \citet{ShimizuDirectLiNGAM2011}.
We also thank Han Bao and Kenshin Abe for proofreading the manuscript.
We would also like to thank Kento Nozawa and Yoshihiro Nagano for maintaining the computational resources used for our experiments.
This work was supported by RIKEN Junior Research Associate Program.
TT was supported by Masason Foundation.
MS was supported by JST CREST Grant Number JPMJCR18A2.
\printbibliography
\clearpage
\appendix
\onecolumn

\global\csname @topnum\endcsname 0
\global\csname @botnum\endcsname 0

\begin{appendices}
Table~\ref{tbl:notation} summarizes the abbreviations and the symbols used in the paper.
For notation simplicity, when \(\ZspjEntire\) is a finite set, we identify it with \(\Int/m\Int\) where \(m\) is the cardinality of \(\ZspjEntire\), to justify the subtractions inside the kernel functions.

\begin{table}[!h]
\caption{Abbreviations and Symbols in the Paper.}
\label{tbl:notation}
\begin{center}
\begin{tabular}{ll}
\multicolumn{1}{c}{\bf ABBREVIATION / SYMBOL}  &\multicolumn{1}{c}{\bf DESCRIPTION} \\
\hline\\
CG/CGM & Causal Graph / Causal Graphical Model \\
ADMG & Acyclic Directed Mixed Graph \\
DAG/PAG & Directed Acyclic Graph / Partial Ancestral Graph \\
MSE & Mean Squared Error \\
\hline
\(\Re,\Renng,\Repos,\Int,\Intnng,\Na\) & Set of all real numbers, nonnegative real numbers, positive real numbers,\\
& integers, nonnegative integers, and positive integers. \\
\(\Indicator{A}\) & Indicator function, i.e., \(1\) if \(A\) holds true and \(0\) otherwise. \\
\(\condIndep{X}{Y}{Z}\) & \(X\) and \(Y\) are conditionally independent given \(Z\). \\
\(\disjointUnion\) & Disjoint union of sets. \\
\(\DiagMatrix{(x_1, \ldots, x_d)}\) & Diagonal matrix with diagonal elements \((x_1, \ldots, x_d) \ (d \in \Na)\). \\
\(\twonrm{\cdot}, \opnrm{\cdot}, \supnrm{\cdot}\), \(\det\) & Euclidean norm of a vector, the operator norm of a matrix,\\
& the supremum norm of a function, and the determinant of a matrix. \\
\(\floor{\cdot}\) & \(\floor{a} := \max\{z \in \mathbb{Z} : z \leq a\}\) for \(a \in \Re\). \\
\(\Dirac{\z}\) & Dirac's delta function centered at \(\z\) (e.g., \citealp[Section~E.4.1]{ZorichMathematical2015}). \\
\(\Simplex{\nRounds}\) & \((K-1)\)-dimensional probability simplex \citep[Example~2.5]{BoydConvex2010}.\\
\(\ntom{N}{M}, [N]\) & \(\ntom{N}{M} := \{N, N+1, \ldots, M\}\) and \([N] := \ntom{1}{N}\), where \(N, M \in \mathbb{N}\) and \(N \leq M\). \\
\(\x^S\) & \(\x^S := (x^{s_1}, \ldots, x^{s_{|S|}})\) where \(\x = (x^1, \ldots, x^n)\) is an \(n\)-dimensional vector and \\
& \(S = \{s_1, \ldots, s_{|S|}\} \subset [n]\) with \(s_1 < \cdots < s_{|S|}\). \\
\([0] = \emptyset, \Re^0 := \{0\}, \x^\emptyset = 0, [N]^0:=\{0\}\) & Conventions used in the paper. \\
\hline
\(\D \in \mathbb{N}\) & Overall data dimensionality (with \(X\) and \(Y\) combined). \\
\(\Zsp = \bigtimes\runj \Zsp_j\) & Overall data space (without distinguishing \(X\) and \(Y\)). \\
\(\Xsp = \bigtimes\runjX\ZspkEntire{j}\), \(\Ysp = \ZspkEntire{\jY}\) & Input variable space and target variable space. \\
\(\p\) & Joint probability density of $\Zs := (\Z^1, \ldots, \Z^\D)$ taking values in \(\Zsp\). \\
\(\Radmq\) & Rademacher complexity of a function class. \\
\(\HypoCls \subset \Ysp^\Xsp\) & Hypothesis set. \\
\(\ell: \HypoCls \times \left(\bigtimes\runj \ZspkEntire{j}\right) \to \Re\) & Loss function. \\
\(\Risk(\f) = \E[\ell(\f, \Zs)]\) & Risk functional for \(\f \in \HypoCls\). \\
\hline
\(\Data = \{\Zs_i\}_{i=1}^n\) & Independently and identically distributed sample from \(\p\). \\
\(\G = ([\D], \Edges, \Biedges)\), \(\hG = ([\D], \hat\Edges, \hat\Biedges)\) & Underlying ADMG for which \(\p\) satisfies the topological ADMG factorization \\
& and its estimator. \\
\(\District{\cdot}, \Parent{\cdot}, \mpj\) & District, parents, and Markov pillow of vertex \(j \in [\D]\). \\
\(\pjmpj, \pjmpjjoint, \pzmpj\) & Conditional density of \(\Zj\) given \(\Zmpj\), the joint density of \((\Zj, \Zmpj)\), \\
& and the marginal density of \(\Zmpj\). \\
\hline
\(\Kj: \ZspmpjEntire \to \Re\) & Kernel function (we define \(\Kj := 1\) if \(\mpi{j} = \emptyset\)). \\
\(\Zs_\bi\) & \(\Zs_\bi = (\Zijk{i_1}{1}, \ldots, \Zijk{i_\D}{\D})\) for \(\bi = (i_1, \ldots, i_\D) \in [n]^\D\). \\
\(\Daug := \{\Zs_\bi\}\runbi, \Waug := \{\wbi\}\runbi\) & Augmented data set and the instance weights. \\
\(\hRemp, \hRaug\) & Ordinary empirical risk estimator and the proposed risk estimator. \\
\(\Reg(\f)\) & Regularization term for \(\f \in \HypoCls\). \\
\(\lam \in [0, 1]\) & Convex combination coefficient used in \((1 - \lam) \hRemp(\f) + \lam\hRaug(\f) + \Reg(\f)\). \\
\(\Kkk{j}{j'}\) & Component of the product kernel \(\Kj\) for \(j' \in \mpj\). \\
\(\weightThreshold\) & Pruning threshold of the small weights in Algorithm~\ref{paper:alg:proposed-method}. \\
\hline
\end{tabular}
\end{center}
\end{table}
\section{Preliminaries on ADMG}
\label{sec:org15ed6f1}
\label{sec:appendix:admg-terminology}
Given an ADMG \(\G\) with the vertex set \(V\) and topological order \(\preceq\), we use the following terminologies~\citep{BhattacharyaSemiparametric2020}.

\paragraph{District.}
For \(v \in V\), define \(\District{v}\) as the collection of \(v' \in V\) that is connected to \(v\) via a bi-directed path.

\paragraph{Parents.}
For a subset \(A \subset V\), we define its parents as \(\Parent{A} := \bigcup_{v \in A} \Parent{v} \setminus A\) where \(\Parent{v}\) denotes the parent of \(v\) in the usual sense.

\paragraph{Markov pillow.}
For \(v \in V\), define \(\G_{\preceq v}\) to be the subgraph of \(\G\) that is composed of only the vertices that precede \(v\).
Then, the Markov pillow of \(v \in V\) is \(\mpi{v} := \District{v} \cup \Parent{\District{v}} \setminus \{v\}\) in \(\G_{\preceq v}\).
Throughout the paper, we use the fact that \(\mpi{v}\) consists only of variables that are precedent to \(v\).
\section{Experiment Details}
\label{sec:orgd3f1a16}
\label{paper:sec:appendix:experiment-detail}
Here, we describe the implementation details of the experiment.
The experiment was implemented using the \emph{hydra} package of Python~\citep{Yadan2019Hydra}.
All experiments were carried out on a \unitnum{2.60}{GHz} Intel\textregistered{}~Xeon\textregistered{} CPUs with \unitnum{132}{GB} memory.

Our experiment code can be found at \small{}\url{https://github.com/takeshi-teshima/incorporating-causal-graphical-prior-knowledge-into-predictive-modeling-via-simple-data-augmentation}.
\subsection{Data Set details}
\label{sec:org3d0b54a}
\label{paper:sec:appendix:dataset}
Following are the data acquisition procedures, the sample sizes, the variable definitions, and the preprocessing procedures used in our experiment.
In all the data sets, after preprocessing as described below, we independently normalized each variable as a final preprocessing step.
\paragraph{Sachs data~\citep{SachsCausal2005}.}
\label{sec:orgf541a4a}
This data set consists of continuous measurements from the flow cytometry of proteins and phospholipids in human immune system cells.
The \emph{consensus graph} is provided in \citet{SachsCausal2005} based on the conventionally accepted cellular signaling networks (Figure~\ref{fig:reference-graph}\subref{fig:sachs-graph}).
Among the eight data sets corresponding to different intervention conditions~\citep{SachsCausal2005}, we use the one that is \emph{observational},
i.e., without any interventions.
The data set contains \(853\) observations of \(11\) variables,
namely \emph{Raf}, \emph{Mek}, \emph{Plcg}, \emph{PIP2}, \emph{PIP3}, \emph{Erk}, \emph{Akt}, \emph{PKA}, \emph{PKC}, \emph{P38}, and \emph{Jnk}.
Among these, for demonstration purposes, we considered \emph{PKA} as the target attribute.
As preprocessing, we log-transformed \emph{Raf}, \emph{Mek}, and \emph{PKA}.
\paragraph{GSS data~\citep{ShimizuDirectLiNGAM2011}.}
\label{sec:org4d65351}
This data set is concerning the status attainment theory in sociology.
This data set is originally part of the General Social Survey (GSS)\footnote{\url{https://gss.norc.org/}},
and we used a subset of the data that was previously used in the causal discovery literature~\citep{ShimizuDirectLiNGAM2011}.
The reference graph is based on domain knowledge of the status attainment model (\citealp{DuncanSocioeconomic1972}; Figure~\ref{fig:reference-graph}\subref{fig:gss-graph}).
The acquired data set consists of \(1380\) observations of \(6\) variables,
namely \emph{\(x_1\)}: father's occupation level, \emph{\(x_2\)}: son's income, \emph{\(x_3\)}: father's education, \emph{\(x_4\)}: son's occupation, \emph{\(x_5\)}: son's education, and \emph{\(x_6\)}: the number of siblings.
We consider \emph{\(x_4\)} as the target variable.
\paragraph{Boston Housing data~\citep{HarrisonHedonic1978}.}
\label{sec:org31e8611}
This data set is concerning the house prices in Boston, and the objective is to predict the prices of the house from its attributes.
We acquired the data from \url{https://github.com/adityatiwari13/Boston\_Dataset}.
The acquired data set consists of \(506\) observations of \(13\) variables,
namely \emph{CRIM}, \emph{ZN}, \emph{INDUS}, \emph{CHAS}, \emph{NOX}, \emph{RM}, \emph{AGE}, \emph{DIS}, \emph{RAD}, \emph{TAX}, \emph{PTRATIO}, \emph{B}, \emph{LSTAT}, and \emph{MEDV}.
The objective is to predict the value of prices of the house, i.e., \emph{MEDV}, using the given features.
\paragraph{Auto MPG data~\citep{QuinlanCombining1993}.}
\label{sec:orgb94a93f}
This data set concerns the city-cycle fuel consumption in miles per gallon (MPG).
We acquired the data from \url{https://archive.ics.uci.edu/ml/datasets/Auto+MPG}.
The acquired data set consists of \(398\) observations of \(9\) variables,
namely \emph{mpg}, \emph{cylinders}, \emph{displacement}, \emph{horsepower}, \emph{weight}, \emph{acceleration}, \emph{model year}, \emph{origin}, and \emph{car name}.
Among these, we discard \emph{origin} and \emph{car name}, and we consider \emph{mpg} as the predicted variable.
\paragraph{White Wine data~\citep{CortezModeling2009}.}
\label{sec:org7b8af6f}
This data set is concerning the prediction of wine quality from its physicochemical attributes.
We acquired the data from \url{https://archive.ics.uci.edu/ml/datasets/wine+quality}.
The acquired data set consists of \(4898\) observations of \(12\) variables,
namely \emph{fixed acidity}, \emph{volatile acidity}, \emph{citric acid}, \emph{residual sugar}, \emph{chlorides}, \emph{free sulfur dioxide}, \emph{total sulfur dioxide}, \emph{density}, \emph{pH}, \emph{sulphates}, \emph{alcohol}, and \emph{quality}.
Among the variables, we consider the \emph{quality} variable as the target.
\paragraph{Red Wine data~\citep{CortezModeling2009}.}
\label{sec:org7b3d1a3}
This data set is concerning the prediction of wine quality from its physicochemical attributes.
We acquired the data from \url{https://archive.ics.uci.edu/ml/datasets/wine+quality}.
The acquired data set consists of \(1599\) observations of \(12\) variables,
namely \emph{fixed acidity}, \emph{volatile acidity}, \emph{citric acid}, \emph{residual sugar}, \emph{chlorides}, \emph{free sulfur dioxide}, \emph{total sulfur dioxide}, \emph{density}, \emph{pH}, \emph{sulphates}, \emph{alcohol}, and \emph{quality}.
Among these, we consider the \emph{quality} variable as the target.
\subsection{Predictor model details}
\label{sec:org7ac4f7b}
\label{paper:sec:appendix:model-details}
For the implementation of the predictor model, we employed the \emph{xgboost} library of Python~\citep{ChenXGBoost2016}.
See \citet{ChenXGBoost2016} for the optimization method and the other details.
\subsection{Proposed method implementation details}
\label{sec:orgf0d2d5f}
\label{paper:sec:appendix:proposed-method-details}
\label{paper:sec:appendix:hyperparameters}
For continuous variables, we compute the kernel bandwidths as follows.
We first specify the \emph{bandwidth temperature} \(\gamma > 0\) as a hyper-parameter.
Then we calculate the rule-of-thumb bandwidth \(\ruleOfThumbBandwidth_j\) for each \(j \in [\D]\) using the training data \(\{\Zs_i^j\}\runi\).
Finally, we set \(h_j = \gamma \cdot \ruleOfThumbBandwidth_j\).
In the experiment, we fix \(\gamma = 10^{-3}\) throughout all runs.

For the rule-of-thumb kernel bandwidth, we employed \emph{Silverman}'s rule-of-thumb~\citep[pp.45--47, Equations~(3.28) and (3.30) therein]{SilvermanDensity1986} implemented in the \emph{statsmodels} package of Python~\citep{statsmodels2020},
namely, \(\ruleOfThumbBandwidth = \left(\frac{4}{3}\right)^{1/5} A n^{-1/5}\)
where \(A = \min\{\hat \sigma, \mathrm{IQR} / 1.349\}\),
\(\hat \sigma\) is the square root of the unbiased estimator of the variance,
and \(\mathrm{IQR}\) is the interquantile range.

For the pruning threshold, we use \(\weightThreshold = 10^{-3} \cdot n^{-1}\).

\subsection{Causal Discovery Method Configuration}
\label{sec:orgfda2e57}
We perform \emph{DirectLiNGAM} \citep{ShimizuDirectLiNGAM2011} on the data sets to simulate a situation where we have access to domain knowledge.
As the independence measure used in the DirectLiNGAM framework, we employ the pairwise likelihood ratio score \citep{HyvarinenPairwise2013} that is based on a nonparametric approximation to the mutual information.
\subsection{Supplementary experiment results}
\label{sec:org74bc4e6}
\label{paper:sec:appendix:experiment-relative-improvement}
\begin{figure*}[t]
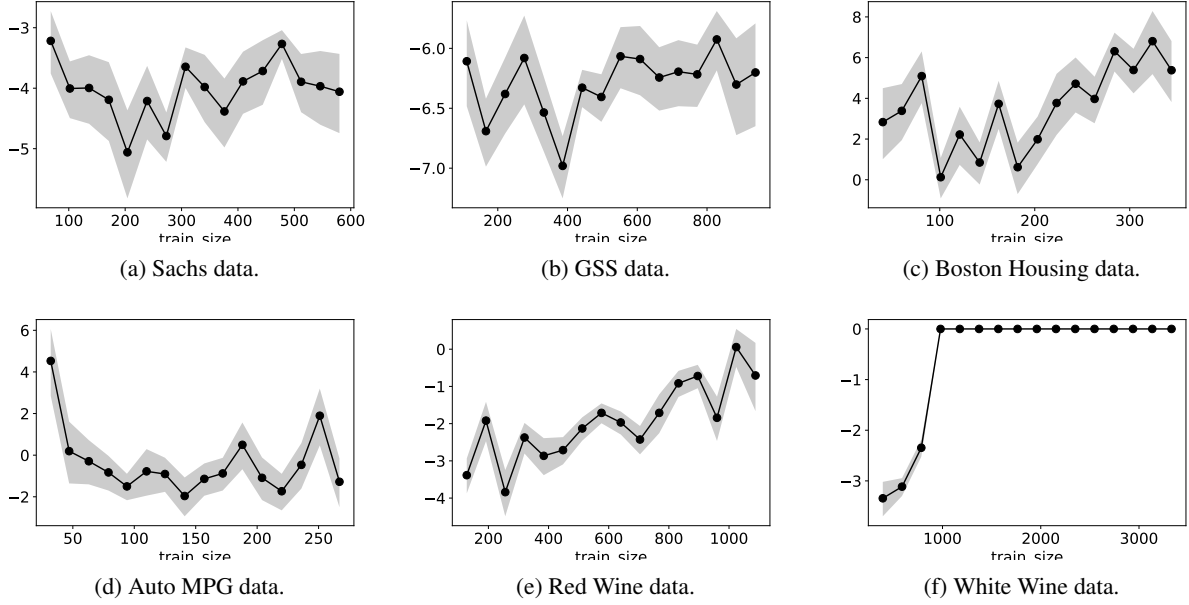

\begin{minipage}[c]{1.0\linewidth}
\begin{minipage}[c]{0.33\linewidth}
\centering{}\DrawSubFig{\figureRoot/relative_improvement_sachs_proposed}{Sachs data.}{fig:rel-sachs}{keepaspectratio, height=\textheight, width=\textwidth}
\end{minipage}\hfill
\begin{minipage}[c]{0.33\linewidth}
\centering{}\DrawSubFig{\figureRoot/relative_improvement_gss_proposed}{GSS data.}{fig:rel-gss}{keepaspectratio, height=\textheight, width=\textwidth}
\end{minipage}\hfill
\begin{minipage}[c]{0.33\linewidth}
\centering{}\DrawSubFig{\figureRoot/relative_improvement_boston_housing_proposed}{Boston Housing data.}{fig:rel-boston_housing}{keepaspectratio, height=\textheight, width=\textwidth}
\end{minipage}\hfill
\end{minipage}

\begin{minipage}[c]{1.0\linewidth}
\begin{minipage}[c]{0.33\linewidth}
\centering{}\DrawSubFig{\figureRoot/relative_improvement_auto_mpg_proposed}{Auto MPG data.}{fig:rel-auto_mpg}{keepaspectratio, height=\textheight, width=\textwidth}
\end{minipage}\hfill
\begin{minipage}[c]{0.33\linewidth}
\centering{}\DrawSubFig{\figureRoot/relative_improvement_red_wine_proposed}{Red Wine data.}{fig:rel-red_wine}{keepaspectratio, height=\textheight, width=\textwidth}
\end{minipage}\hfill
\begin{minipage}[c]{0.33\linewidth}
\centering{}\DrawSubFig{\figureRoot/relative_improvement_white_wine_proposed}{White Wine data.}{fig:rel-white_wine}{keepaspectratio, height=\textheight, width=\textwidth}
\end{minipage}\hfill
\end{minipage}

\begin{minipage}[c]{1.0\linewidth}
\caption{
Average relative improvement in percentage.
In all figures, the horizontal axis is the varied sizes of the original training data before augmentation.
The vertical axis is the relative MSE improvement in percentage, i.e., \(\frac{\MSEProposed - \MSEBaseline}{\MSEBaseline} \times 100\) \%
where \(\MSEBaseline\) and \(\MSEProposed\) are the MSE of the baseline and that of the proposed method, respectively (the lower the better).
The markers and the lines indicate the average over the 20 independent runs, and the shades are drawn for the width of the standard errors both above and below the lines.
In most of the cases, the proposed method shows a consistently improved performance compared to the baseline based on the empirical risk minimization with the same hypothesis class, particularly in the small-data regime.
}
\label{fig:appendix-result-1}
\end{minipage}
\end{figure*}

Figure~\ref{fig:appendix-result-1} shows the average improvement achieved by the proposed method relative to the baseline without a device.
The improvement in the small-data regime is consistently observed except in a few cases in the \emph{Auto MPG} and the \emph{Boston Housing} data.
In the \emph{Boston Housing} data set, the performance loss may be due to the failure of the CG estimation since the performance loss is magnified as the training set size is increased.
In the \emph{Auto MPG} data, the performance degradation for the smallest training set fraction may be due to the additional complexity and bias introduced by the kernel approximation.
\section{Details and Proof of the Theoretical Analysis}
\label{sec:orgb8f80c9}
\subsection{Notation and Problem Setup}
\label{sec:orgf79ef9b}
\paragraph{Basic notation.}
\label{sec:org1cf92bb}
Let \(\Re\) denote the set of real numbers,
\(\Na\) that of positive integers,
\(\Repos\) that of positive real numbers,
\(\Int\) that of integers,
and \(\Intnng\) that of non-negative integers.
For \((x_1, \ldots, x_k) \in \Re^k\), \(\DiagMatrix{(x_1, \ldots, x_k)}\) denotes the diagonal matrix whose diagonal elements are \((x_1, \ldots, x_k)\).
For a vector, \(\twonrm{\cdot}\) denotes its Euclidean norm.
For a matrix, \(\det\) denotes its determinant, and \(\opnrm{\cdot}\) its operator norm.
For a function, \(\supnrm{\cdot}\) denotes its supremum norm over a suitable set of inputs when the domain is clear from the context.
For a finite set, \(\card{\cdot}\) denotes its cardinality.
\paragraph{Utility notation.}
\label{sec:org58a669d}
For \(n \in \Na\), define \([n] := \{1, 2, \ldots, n\}\).
For \(n, m \in \Na\) with \(n \leq m\), define \(\ntom{n}{m} := \{n, n+1, \ldots, m\}\).
For an \(n\)-dimensional vector \(\x = (x_1, \ldots, x_n)\) and \(S \subset [n]\), we let \(\x^S = (x_{s_1, \ldots, s_{|S|}})\) denote its sub-vector with indices in \(S = \{s_1, \ldots, s_{|S|}\}\) with \(s_1 < \cdots < s_{|S|}\). Similarly, for \(j \in [n]\), we let \(\x^j := \x^{\{j\}}\).
For \(S \subset [n]\), we also define \(\Zspk{S} := \bigtimes_{k \in S}\Zspk{k}\).
To simplify the notation, we use the convention of \(\Re^0 := \{0\}\), \(\x^\emptyset = 0\), and \([n]^{j-1} = \{0\}\).
\paragraph{Distribution and sample.}
\label{sec:org010afc0}
Let \(\D \in \Na\).
In this theoretical analysis, we assume that \(\Zspj\) is a measurable subset of \(\Re \ (j \in [\D])\).
We consider a probability distribution over \(\ZspAll := \bigtimes\runj\Zspj\), and let \(\p\) denote its density function (assuming it exists).
We are given \(\Data = \{\Zs_i\}_{i=1}^n\), an independently and identically distributed sample from \(\p\).
Let \(\E\) denote the expectation with respect to \(\p\).
Additionally, we are given an ADMG \(\G = ([\D], \Edges, \Biedges)\).
Let \(\mpi{j} \subset [\D]\) denote the Markov pillow of \(j \in [\D]\).
Throughout this section, we assume \(\p\) satisfies the topological ADMG factorization relation according to \(\G\)~\citep{BhattacharyaSemiparametric2020}:
\begin{align*}
\p(\z) = \prod\runj \pjmpj(\zj | \zmp)
\quad \left(= \prod\runj \frac{\pjmpjjoint(\zj, \zmp)}{\pzmpj(\zmp)}\right).
\end{align*}
\paragraph{Learning problem.}
\label{sec:org479bc79}
Let \(\HypoCls\) denote a hypothesis class, and let \(\ell: \HypoCls \times \ReAll \to \Repos\) be a loss function.
To simplify the notation, we define \(\lf := \ell(\f, \cdot)\) and \(\lF := \{\lf : f \in\HypoCls\}\).
For each \(\f \in \HypoCls\), we define the risk functional \(\Risk(\f) := \E[\lf(\Zs)]\).
The learning problem is to find a hypothesis \(\hf \in \HypoCls\) for which \(\Risk\) is small, given the training data \(\Data\) and the graph \(\G\).
\paragraph{Proposed method.}
\label{sec:org559c679}
For each \(j \in [\D]\), we fix a kernel function \(\Kj : \Rempj \to \Re\).
For notation simplicity, we define \(\Kj := 1\) for \(j\) such that \(\mpi{j} = \emptyset\).
We also fix \(\h = (\hk{1}, \ldots, \hk{\D}) \in \Repos^\D\).
Then, we define
\begin{align*}
\Hj := \DiagMatrix{\hk{\mpj}}, \qquad
\KHj(u) := \frac{1}{|\det\Hj|}\Kj(\Hj^{-1} u).
\end{align*}
For \(\bi = (i_1, \ldots, i_\D)\) and \(\zmp \in \Rempj\), define
\begin{align*}
\wij(\zmp) := \frac{\KHj(\zmp-\Zmpi)}{\sum\runi \KHj(\zmp-\Zmpi)}\Indicator{\sum\runi \KHj(\zmp-\Zmpi)\neq0}
\end{align*}
where \(\bi = (i_1, \ldots, i_\D)\), \(\zmp \in \Rempj\).
Then, we recursively define
\begin{align*}
\wiii{\bikk{1}{0}} = 1,\quad \wiiij{\bij}{j} = \wseqij{\bijm}{i_j}{j} \cdot \wiiij{\bijm}{j-1} \ (j \in [\D], \bijm \in [n]^{j-1}),
\end{align*}
where
\begin{align*}
\wbij := \wijj\left(\Zbijmpj\right), \quad \Zbij= \left(\Zijk{i_1}{1}, \ldots, \Zijk{i_{j-1}}{j-1}\right).
\end{align*}
Here, we use the convention \(\Zijk{\bikk{1}{0}}{\mpi{1}} := 0\) to be consistent with the notation.
Using this notation, for \(\f \in \HypoCls\), define the augmented empirical risk estimator
\begin{align*}
\hRaug(\f) := \sum\runbi \wbi \lf(\Zs_\bi).
\end{align*}
\paragraph{Target of the theoretical analysis.}
\label{sec:orga781a68}
We aim to provide a stochastic upper bound on \(\Risk(\hf) - \Risk(\fstar)\),
where
\begin{align*}
\hf \in \argmin_{\f \in \HypoCls} \{\hRaug(\f)\}, \text{ and } \fstar \in \argmin\runf \{\Risk(\f)\},
\end{align*}
assuming both exist.
\paragraph{Notation for stating the results.}
\label{sec:orga4256f3}
To state the main theorem, we use the following notation.
For each \(j \in [\D]\) and \(\f \in \HypoCls\), define
\begin{align*}
\lfj &: \bivec{\z^1}{\zj} \mapsto \int_\Zspkk{j+1}{\D} \lf(\z)\left(\prod\runfromj{k} \pkmpk{k}(\zk{k} | \zmpk{k})\right)\dz^{j+1} \cdots \dz^{\D}.
\end{align*}
Also define
\begin{align*}
\lFjOne &:= \left\{\lfj(\zk{1}, \ldots, \zk{j-1}, \cdot): \f \in \HypoCls, (\zk{1}, \ldots, \zk{j-1})\in\Zspkk{1}{j-1}\right\}, \\
\KernelShiftedClassHj &:= \left\{\KHj(\zmp-(\cdot)): \setzmp\right\}.
\end{align*}

For simplicity, throughout the theoretical analysis, we assume that all quantities appearing in the proof satisfy sufficient measurability conditions.

\subsection{Main Theorem}
\label{sec:orga682568}
\label{paper:sec:appendix:theory-1}
Here, we detail the assumptions, the statement, and a proof of Theorem~\ref{paper:thm:1}.
\subsubsection{Preliminaries}
We use the following convenient \emph{multi-index} notation (see, e.g., \citealp{StoneOptimal1982}).
\begin{definition}[Multi-index notation]
For \(d \in \Na\), we call a \(d\)-tuple \(\alpha = (\alpha_1, \ldots, \alpha_d) \in \MultiIndexSpace{d}\) \emph{multi-index}.
For a multi-index \(\alpha\), let \(|\alpha| := \sum_{j=1}^d \alpha_j\) and \(\alpha! := \prod_{j=1}^d \alpha_j!\), and \(x^\alpha = x_1^{\alpha_1} \cdots x_d^{\alpha_d}\) for \(x = (x_1, \ldots, x_d) \in \Re^d\).
Also, let \(\Deriv{\alpha}{}\) denote the partial differential operator defined by
\begin{align*}
\Deriv{\alpha}{} = \frac{\partial^{|\alpha|}}{\partial x_1^{\alpha_1} \cdots \partial x_d^{\alpha_d}}.
\end{align*}
\end{definition}
\begin{definition}[Convolution]
Let \(d \in \Na\) and \(\Omega \subset\Re^d\) be a measurable subset.
For continuous bounded functions \(f, g: \Omega \to \Re\), we define a function \((\convolutionFull{f}{g}{\Omega}) : \Omega \to \Re\) by
\begin{align*}
\convolutionFull{f}{g}{\Omega}(\x) := \int_\Omega f(\x - \y) g(\y) \dr\y.
\end{align*}
When \(\Omega = \Re^d\), we drop \(\Omega\) from the notation and denote \(\convolution{f}{g}\).
\end{definition}

We define the following class of functions.
\begin{definition}[\HolderName{} class; \citealp{StoneOptimal1982,TsybakovIntroduction2009}]
Let \(d \in \Na\), \(\beta > 1\), \(L > 0\), and let \(\Omega \subset \Re^d\) be an open subset.
The \((\beta, L)\)-\HolderName{} class \(\HolderClassFull{\beta}{L}{\Omega}\) is defined as the set of \(k = \floor{\beta}\)-times continuously differentiable functions \(f: \Omega \to \Re\) satisfying
\begin{align*}
|\Deriv{\alpha}{f}(x) - \Deriv{\alpha}{f}(x')| \leq L \twonrm{x - x'}^{\beta - k} \quad\text{for}\quad x, x' \in \Omega \text{ and } |\alpha| = k,
\end{align*}
where \(\alpha = (\alpha_1, \ldots, \alpha_d) \in \MultiIndexSpace{d}\) is a multi-index, and \(\floor{a} = \max\{z \in \mathbb{Z} : z \leq a\}\) for \(a \in \Re\).
When \(\Omega=\Re^d\), we also drop \(\Re^d\) from the notation and denote \(\HolderClass{\beta}{L}\) when the dimension is clear from the context.
\end{definition}

\begin{remark}
In the 1-dimensional case, a related analysis based on the notion of the \HolderName{} class is presented in Section~1.2.3 of \citet{TsybakovIntroduction2009}.
\end{remark}
For function classes, we quantify their complexities using the Rademacher complexity.
\begin{definition}[Rademacher complexity]
Let \(q\) denote a probability distribution on some measurable space \(\mathcal{X}\).
For a function class \(\mathcal{F} \subset \Re^\mathcal{X}\), define
\begin{align*}
\Radmq(\mathcal{F}) := \E_q\Erad \left[\sup_{f \in \mathcal{F}} \left|\frac{1}{m} \sum_{i=1}^m \rad_i f(X_i)\right|\right]
\end{align*}
where \(m \in \Na\), \(\{\rad_i\}_{i=1}^m\) are independent uniform \(\{\pm 1\}\)-valued random variables, and \(\{X_i\}_{i=1}^m \overset{\text{i.i.d.}}{\sim} q\).
\end{definition}
\subsubsection{Assumptions}
 For simplicity, throughout this theoretical analysis, we assume that all quantities appearing in the proof satisfy sufficient measurability conditions.
\begin{assumption}[Boundedness assumptions]
We assume that the following hold:
\begin{itemize}
\item The loss function is bounded, i.e., \(\lossBound := \supf\sup_{\Zs \in \ReAll}|\ell(\f, \Zs)| < \infty\).
\item \(\KAll := \{\Kj\}\runj\) are uniformly bounded from above, i.e., \(\keBound := \max\left\{\supnrm{\Kj}: j \in [\D]\right\} < \infty\).
\item For each \(j \in [\D]\), \(\Zspj \subset \Re\) is a compact subset. Let \(\ZspjBound := \int_{\Zspj}\dzj < \infty\).
\item For all \(j \in [\D]\), \(\pzmpj\) is bounded away from zero over \(\Asp\). Define \(\epsmpj := \inf_{\zmp \in \Zspmpj}\pzmpj(\zmp)\).
\item For each \(j \in [\D]\), \(\Kj\) is continuous and strictly positive.
We define
\begin{align*}
\expandedDiffKHj := \sup_{\stackrel{\zmp \in \Zspmpj,}{\zmpp \in \Rempj\setminus\Zspmpj}} \left|\KHj(\zmp - \zmpp)\right|
= \sup_{\stackrel{\zmp \in \invHj\Zspmpj,}{\zmpp \in \invHj(\Rempj\setminus\Zspmpj)}} \left|\Kj(\zmp - \zmpp)\right|\invdetHj
\end{align*}
and assume \(\expandedDiffKHj < \infty\).
\end{itemize}
\label{paper:assumption:boundedness}
\end{assumption}

\begin{remark}[]
Since \(\Zspmpj\) is compact and \(\Kj\) is continuous,
if we define
\begin{align*}
\epsKHj &:= \absdet{\Hj} \left(\inf_{\x, \x' \in \Zspmpj} \KHj(\x - \x')\right) = \inf_{\x, \x' \in \Hj^{-1}\Zspmpj} \Kj(\x - \x'),
\end{align*}
this quantity is strictly positive under Assumption~\ref{paper:assumption:boundedness}.
\end{remark}
From here, we fix \(\beta > 1\) and \(L > 0\).
\begin{assumption}[Smoothness assumptions]
We assume that the following hold for all \(j \in [\D]\):
\begin{itemize}
\item \(\pzmpj\) has an extension \(\pzmpjExt \in \HolderClass{\beta}{L}\) such that \(\pzmpjExtInt := \int_{\Rempj\setminus\Zspmpj}|\pzmpjExt(\zmp)|\dzmp < \infty\).
\item For all \(\zj \in \Zspj\), \(\pjmpjjoint(\zj, \cdot)\) has an extension \(\pjmpjjointExt(\zj, \cdot) \in \HolderClass{\beta}{L}\) such that \(\pjmpjjointExtInt := \int_{\Zspj}\left(\int_{\Rempj\setminus\Zspmpj}|\pjmpjjointExt(\zj, \zmp)|\dzmp\right)\dzj < \infty\).
\item \(\Kj\) is of order \(k = \floor{\beta}\), i.e.,
\begin{align*}
&\int_{\Rempj} \Kj(u) \du = 1,\qquad
\int_{\Rempj} \Kj(u) u^\alpha \du = 0 \quad (1 \leq |\alpha| \leq k),
\end{align*}
where \(\alpha\in \MultiIndexSpace{\mpjsize}\) is a multi-index,
and \(\Kj\) satisfies \(\int_{\Rempj} |\Kj(u)| \cdot \twonrm{u}^{\beta}\du < \infty\).
\end{itemize}
\label{paper:assumption:smoothness}
\end{assumption}

\begin{remark}[Existence of the smooth extensions]
The smooth extensions in Assumption~\ref{paper:assumption:smoothness} exist, for example,
if we consider a smooth density function \(\pzmpjExt\) on \(\Rempj\)
and regard its restriction to \(\Zspmpj\) with appropriate scaling as \(\pzmpj\).
\end{remark}
\subsubsection{Statement and Proof}
We prove the following theorem.
Theorem~\ref{paper:thm:1} is obtained by changing \(\delta\) to \(\frac{\delta}{2\D}\) in the following theorem,
substituting \(\opnrm{\Hj} = \max_{j' \in \mpj}\hk{j'}\), and defining the appropriate constants.
\begin{theorem}[Excess risk bound]
Assume that Assumptions~\ref{paper:assumption:boundedness} and \ref{paper:assumption:smoothness} hold.
Let \(n \in \Na\).
For \(j \in [\D]\), define
\begin{align*}
&\ConstSumHj := \lossBound \sum\runj \frac{1}{\epsmpj}\left(\ZspjBound + \frac{\keBound}{\epsKHj}\right) \HolderBoundConst{\beta}{L}{\Kj} \opnrm{\Hj}^\beta,
\quad\ConstSumIj := \lossBound\sum\runj \frac{\expandedDiffKHj}{\epsmpj} \left(\pjmpjjointExtInt + \frac{\keBound}{\epsKHj}\pzmpjExtInt\right), \\
&\ConstMaxEpsK := \max_{j \in [\D]} \left\{\frac{1}{\epsKHj}, \frac{\keBound}{(\epsKHj)^2}\right\},
\quad\RadHypoK := \sum\runj \absdet{\Hj}\lFjOneKRad,
\quad\RadK := \sum\runj \absdet{\Hj}\KRad.
\end{align*}
Then, for any \(\delta \in (0, 1)\), we have with probability at least \(1 - 2\D\delta\),
\begin{align*}
\Risk(\hf) - \Risk(\fstar)
\leq 2(\ConstSumHj + \ConstSumIj)
+ 4\ConstMaxEpsK(\RadHypoK + \lossBound\RadK)
+ 2D\lossBound\keBound\ConstMaxEpsK\sqrt{\frac{\log(2/\delta)}{2n}}.
\end{align*}
\label{paper:thm:1:restatement}
\end{theorem}
\paragraph{Proof overview.}
 Our proof derives ideas from the literature on \emph{local empirical processes} and \emph{kernel-type estimators}, namely \citet{EinmahlEmpirical2000,EinmahlUniform2005,DonyUniform2006}.
 Two elementary calculations are essential in the proof.
 The first one handles a difference between two products:
 let \(N \in \Na\), \((a_1, \ldots, a_N) \in \Re^N\), and \((b_1, \ldots, b_N) \in \Re^N\), then,
\begin{align}\label{eq:product-difference}
\left(\prod_{j=1}^N a_i\right) - \left(\prod_{j=1}^N b_i\right) = \sum_{j=1}^N a_1 \cdots a_{j-1} (a_j - b_j) b_{j+1} \cdots b_N.
\end{align}
The second one bounds a difference between two ratios from above:
for \(A, B, C, D \in \Re\) with \(B, D \neq 0\),
\begin{align}\label{eq:ratio-difference-bound}
\left|\frac{A}{B} - \frac{C}{D}\right| &= \left|\frac{A}{B} - \frac{C}{B} + \frac{C}{B} - \frac{C}{D}\right| \leq \left|\frac{1}{B}\right| \cdot |A - C| + \left|\frac{C}{BD}\right| \cdot |B - D|.
\end{align}
\begin{proof}[Proof of Theorem~\ref{paper:thm:1:restatement}]
First, note
\begin{align*}
\Risk(\hf) - \Risk(\fstar) &= \Risk(\hf) - \hRa(\hf) + \hRa(\hf) - \Risk(\fstar)
\leq \Risk(\hf) - \hRa(\hf) + \hRa(\fstar) - \Risk(\fstar)
\leq 2 \annot{\supf |\Risk(\f) - \hRa(\f)|}{(*)}.
\end{align*}
For ease of notation, define
\(\hpj(\zj | \zmp) = \sum\runi \Dirac{\Zij}(\zj)\wij(\zmp)\)
and temporarily denote \(\pk{k} := \pkmpk{k}\).
With this notation, \(\hRa(\f) = \int_\ZspAll \lf(\z)\prod\runj \hpj(\zj | \zmp) \dz\).
Then, applying the argument of \Equation{eq:product-difference}, we have
\begin{align*}
\text{(*)}
&= \supf \left|\int_\ZspAll \lf(\z)\prod\runj\pj(\zj | \zmp)\dz - \int_\ZspAll \lf(\z)\prod\runj\hpj(\zj | \zmp)\dz \right| \\
&= \supf \left|\int_\ZspAll \lf(\z)\sum\runj \left(\prod\runfromj{k} \pk{k}(\zk{k} | \zmpk{k})\right) (\pj(\zj | \zmp) - \hpj(\zj | \zmp)) \left(\prod\runtoj{k} \hpk{k}(\zk{k} | \zmpk{k})\right)\dz \right| \\
&\leq \sum\runj \annot{\supf \left|\int_\ZspAll \lf(\z) \left(\prod\runfromj{k} \pk{k}(\zk{k} | \zmpk{k})\right) (\pj(\zj | \zmp) - \hpj(\zj | \zmp)) \left(\prod\runtoj{k} \hpk{k}(\zk{k} | \zmpk{k})\right)\dz \right|}{(*\(j\))}.
\end{align*}
Now, for \(\f \in \HypoCls\) and \(j \in [\D]\), we define \(\lfjbi : \zj \mapsto \lfj(\Zbij, \zj)\).
Then, for each \(j \in [D]\), applying Lemma~\ref{paper:lem:weights}, we obtain
\begin{align*}
\text{(*\(j\))}
&= \supf \left|\sum\runn{i_1}\cdots\sum\runn{i_{j-1}} \left(\int_{\Zspj} \lfjbi(\zj) \pj(\zj | \Zmpij) \dzj - \sum\runn{i_j}\lfjbi(\Z_{i_j}^j) \wbij \right) \wbijm \cdots \wbione\right| \\
&\leq 1 \cdot \left(\supf \max\runbij\left|\int_{\Zspj} \lfjbi(\zj) \pj(\zj | \Zmpij) \dzj - \sum\runn{i_j}\lfjbi(\Z_{i_j}^j) \wijj(\Zbij^\mpi{j})\right|\right) \\
&\leq \max\runbij\supf\sup\runzmp\left|\int_{\Zspj} \lfjbi(\zj) \pj(\zj | \zmp) \dzj - \sum\runn{i_j}\lfjbi(\Z_{i_j}^j) \wijj(\zmp)\right| \\
&\leq \sup\runlfjOne\sup\runzmp \annot{\left|\int_{\Zspj} \lfj'(\zj) \pj(\zj | \zmp) \dzj - \sum\runn{i_j}\lfj'(\Z_{i_j}^j) \wijj(\zmp)\right|}{(**)},
\end{align*}
where we used that \(\left\{\Zmpij\right\}_{\setbij}\subset \Zspmpj\) that follows from \(\left\{\Zimpj\right\}\runi \subset \Zspmpj\).
Define
\begin{align*}
\rj(f, \zmp) &:= \int_{\Zspj} f(\zj)\pjmpjjoint(\zj, \zmp) \dzj,&
\hrj(f, \zmp) &:= \frac{1}{n}\sum\runi f(\Zij)\KHj(\zmp-\Zmpi),\\
\gj(\zmp) &:= \pzmpj(\zmp),&
\hgj(\zmp) &:= \frac{1}{n}\sum\runi \KHj(\zmp-\Zmpi).
\end{align*}
Then, for each \(\setlfjOne\) and \(\setA\),
\begin{align*}
\text{(**)} &= \left|\frac{\rj(\lfj', \zmp)}{\gj(\zmp)} - \frac{\hrj(\lfj', \zmp)}{\hgj(\zmp)}\right| \\
&\leq \annot{\left|\frac{\rj(\lfj', \zmp)}{\gj(\zmp)} - \frac{\E\hrj(\lfj', \zmp)}{\E\hgj(\zmp)}\right|}{\(\rho_1\)} + \annot{\left|\frac{\E\hrj(\lfj', \zmp)}{\E\hgj(\zmp)} - \frac{\hrj(\lfj', \zmp)}{\hgj(\zmp)}\right|}{\(\rho_2\)}.
\end{align*}
By applying the argument of \Equation{eq:ratio-difference-bound}, we can bound each ratio difference term as
\begin{align*}
&\rho_1 \leq \left|\frac{1}{\gj(\zmp)}\right|\cdot |\rj(\lfj', \zmp) - \E\hrj(\lfj', \zmp)| + \left|\frac{\E\hrj(\zmp)}{\gj(\zmp)\E\hgj(\zmp)}\right|\cdot|\gj(\zmp) - \E\hgj(\zmp)| \\
&\rho_2 \leq \left|\frac{1}{\E\hgj(\zmp)}\right|\cdot|\E\hrj(\lfj', \zmp) - \hrj(\lfj', \zmp)| + \left|\frac{\hrj(\zmp)}{\E\hgj(\zmp)\hgj(\zmp)}\right|\cdot|\E\hgj(\zmp) - \hgj(\zmp)|.
\end{align*}
Applying Lemma~\ref{paper:lem:coefficients} to the coefficients,
Lemma~\ref{paper:lem:bias} to the deterministic difference terms bounding \(\rho_1\),
Lemma~\ref{paper:lem:probabilistic} to the stochastic difference terms bounding \(\rho_2\) along with the union bound,
for any \(\delta \in (0, 1)\),
we have with probability at least \(1 - 2D\delta\),
\begin{align*}
\Risk(\hf) - \Risk(\fstar)
\leq 2 \sum\runj &\biggl(\BOne \left(\BiasR\right) \\
                         &\quad  + \BOne\cdot\BTwo \left(\BiasG\right)\\
                         &\quad  + \BThree \left(\ProbR\right)\\
                         &\quad  + \BThree\cdot\BFour \left(\ProbG\right) \biggr).
\end{align*}
By reorganizing the terms, we obtain the assertion.
\end{proof}
\subsubsection{Lemmas}
 Here, we prove the lemmas used in the proof of Theorem~\ref{paper:thm:1:restatement}.
\begin{lemma}[Bounded coefficients]
Assume Assumption~\ref{paper:assumption:boundedness} holds.
Let \(j \in [\D]\).
Then,
\begin{align*}
\sup\runA &\left|\frac{1}{\gj(\zmp)}\right| \leq \BOne,&
\sup\runlfjOne\sup\runA &\left|\frac{\E\hrj(\lfj', \zmp)}{\E\hgj(\zmp)}\right| \leq \BTwo, \\
\sup\runA &\left|\frac{1}{\E\hgj(\zmp)}\right| \leq \BThree,&
\sup\runlfjOne\sup\runA &\left|\frac{\hrj(\lfj', \zmp)}{\hgj(\zmp)}\right| \leq \BFour.
\end{align*}
\label{paper:lem:coefficients}
\end{lemma}
\begin{proof}
By Assumption~\ref{paper:assumption:boundedness}, we have
\begin{align*}
\sup\runA \left|\frac{1}{\gj(\zmp)}\right| &= \frac{1}{\inf\runA \pzmpj(\zmp)} \leq \BOne.
\end{align*}
Also,
\begin{align*}
\sup\runA \left|\frac{1}{\E\hgj(\zmp)}\right|
&\leq \frac{1}{\inf\runA \left|\E\hgj(\zmp)\right|}\\
&= \frac{1}{\inf\runA \left|\int_{\Asp} \KHj(\zmp - \zmpp) \gj(\zmpp)\dr{\zmpp}\right|}\\
&= \frac{1}{\inf\runA \int_{\Asp} \KHj(\zmp - \zmpp) \gj(\zmpp)\dr{\zmpp}}\\
&\leq \frac{1}{\invdetHj\epsKHj \int_{\Asp}\gj(\zmpp)\dr{\zmpp}}
= \BThree,
\end{align*}
where we used the positivity of the integrand.
Now,
\begin{align*}
&\sup\runlfjOne\sup\runA \left|\frac{\E\hrj(\lfj', \zmp)}{\E\hgj(\zmp)}\right|
= \sup\runlfjOne\sup\runA \left|\frac{\absdet{\Hj}\E\hrj(\lfj', \zmp)}{\absdet{\Hj}\E\hgj(\zmp)}\right|\\
&\quad\leq \frac{\sup\runlfjOne\sup\runA \supnrm{\lfj'} \cdot \supnrm{\left(\absdet{\Hj} \KHj\right)}}{\inf\runA\absdet{\Hj}|\E\hgj(\zmp)|}
\leq \BTwo.
\end{align*}
Similarly, we have \(\inf\runA \absdet{\Hj}\cdot\left|\hgj(\zmp)\right| \geq \epsKHj\). Therefore,
\begin{align*}
&\sup\runlfjOne\sup\runA \left|\frac{\hrj(\lfj', \zmp)}{\hgj(\zmp)}\right|
= \sup\runlfjOne\sup\runA \left|\frac{\absdet{\Hj}\hrj(\lfj', \zmp)}{\absdet{\Hj}\hgj(\zmp)}\right|\\
&\quad\leq \frac{\sup\runlfjOne\sup\runA\absdet{\Hj}\cdot\left|\hrj(\lfj', \zmp)\right|}{\inf\runA \absdet{\Hj}\cdot\left|\hgj(\zmp)\right|}
\leq \BFour.
\end{align*}
\end{proof}
\begin{lemma}[Deterministic terms]
Assume that Assumptions~\ref{paper:assumption:boundedness} and \ref{paper:assumption:smoothness} hold.
Let \(j \in [\D]\).
Then,
\begin{align*}
\sup\runlfjOne\sup\runzmp |\rj(\lfj', \zmp) - \E\hrj(\lfj', \zmp)| &\leq \BiasR,\\
\sup\runzmp|\gj(\zmp) - \E\hgj(\zmp)| &\leq \BiasG.
\end{align*}
\label{paper:lem:bias}
\end{lemma}
\begin{proof}
By applying Lemma~\ref{lem:Holder-conv-bound} under Assumption~\ref{paper:assumption:smoothness},
\begin{align*}
&\sup\runzmp |\gj(\zmp) - \E\hgj(\zmp)| \\
&= \sup\runzmp \left|\pzmpj(\zmp) - \int_{\Zspmpj} \KHj(\zmp-\zmpp)\pzmpj(\zmpp)\dr\zmpp\right| \\
&= \sup\runzmp \left|\pzmpjExt(\zmp) - \int_{\Zspmpj} \KHj(\zmp-\zmpp)\pzmpjExt(\zmpp)\dr\zmpp\right| \\
&\leq \sup\runzmp \left|\pzmpjExt(\zmp) - \left(\convolution{\KHj}{\pzmpjExt}\right)(\zmp)\right|
+ \sup\runzmp \left|\int_{\Rempj\setminus\Zspmpj} \KHj(\zmp-\zmpp)\pzmpjExt(\zmpp)\dr\zmpp\right|\\
&\leq \HolderBoundConst{\beta}{L}{\Kj} \opnrm{\Hj}^\beta + \expandedDiffKHj\pzmpjExtInt.
\end{align*}
Similarly, for each \(\setlfjOne\) and \(\setzmp\),
\begin{align*}
&|\rj(\lfj', \zmp) - \E\hrj(\lfj', \zmp)| \\
&= \left|\int_{\Zspj} \lfj'(\zj)\pjmpjjoint(\zj, \zmp)\dzj - \int_{\Zspj} \lfj'(\zj) \left(\int_{\Zspmpj} \KHj(\zmp-\zmpp)\pjmpjjoint(\zj, \zmpp)\dr\zmpp\right)\dzj\right| \\
&= \left|\int_{\Zspj} \lfj'(\zj)\pjmpjjointExt(\zj, \zmp)\dzj - \int_{\Zspj} \lfj'(\zj) \left(\int_{\Zspmpj} \KHj(\zmp-\zmpp)\pjmpjjointExt(\zj, \zmpp)\dr\zmpp\right)\dzj\right| \\
&\leq \left|\int_{\Zspj} \lfj'(\zj)\left(\pjmpjjointExt(\zj, \zmp) - (\convolution{\KHj}{\pjmpjjointExt(\zj, \cdot)})(\zmp)\right)\dzj\right|\\
&\quad+ \left|\int_{\Zspj} \lfj'(\zj) \left(\int_{\Rempj\setminus\Zspmpj} \KHj(\zmp-\zmpp)\pjmpjjointExt(\zj, \zmpp)\dr\zmpp\right)\dzj\right|\\
&\leq \lossBound \int_{\Zspj} \left|\pjmpjjointExt(\zj, \zmp) - (\convolution{\KHj}{\pjmpjjointExt(\zj, \cdot)})(\zmp)\right|\dzj + \lossBound\expandedDiffKHj\pjmpjjointExtInt\\
&\leq \lossBound \ZspjBound \sup\runzj\left|\pjmpjjointExt(\zj, \zmp) - (\convolution{\KHj}{\pjmpjjointExt(\zj, \cdot)})(\zmp)\right| + \lossBound\expandedDiffKHj\pjmpjjointExtInt \\
&\leq \lossBound \ZspjBound \sup\runzj\sup\runzmp\left|\pjmpjjointExt(\zj, \zmp) - (\convolution{\KHj}{\pjmpjjointExt(\zj, \cdot)})(\zmp)\right| + \lossBound\expandedDiffKHj\pjmpjjointExtInt.
\end{align*}
Applying Lemma~\ref{lem:Holder-conv-bound} under Assumption~ \ref{paper:assumption:smoothness},
for each \(\setzj\), we obtain
\begin{align*}
\sup\runzmp\left|\pjmpjjointExt(\zj, \zmp) - (\convolution{\KHj}{\pjmpjjointExt(\zj, \cdot)})(\zmp)\right| \leq \HolderBoundConst{\beta}{L}{\Kj} \opnrm{\Hj}^\beta.
\end{align*}
Therefore, we have the assertion.
\end{proof}
\begin{lemma}[Probabilistic terms]
Assume that Assumption~\ref{paper:assumption:boundedness} holds.
Let \(j \in [\D]\).
For any \(\delta \in (0, 1)\), with probability at least \(1 - \delta\), we have
\begin{align*}
\sup\runlfjOne\sup\runzmp&|\E\hrj(\lfj', \zmp) - \hrj(\lfj', \zmp)|
\leq \ProbR.
\end{align*}
Similarly, for any \(\delta \in (0, 1)\), with probability at least \(1 - \delta\), we have
\begin{align*}
\sup\runzmp|\E\hgj(\zmp) - \hgj(\zmp)| \leq \ProbG.
\end{align*}
\label{paper:lem:probabilistic}
\end{lemma}
\begin{proof}
Note
\begin{align*}
&\sup\runlfjOne\sup\runzmp|\E\hrj(\lfj', \zmp) - \hrj(\lfj', \zmp)|
= \sup\runlfjOne\sup\runk\left|\frac{1}{n}\sum\runi \lfj'(\Zij)\ke(\Zmpi) - \E\left[\frac{1}{n}\sum\runi \lfj'(\Zij)\ke(\Zmpi)\right]\right|
\end{align*}
and
\begin{align*}
\sup\runzmp |\E\hgj(\zmp) - \hgj(\zmp)|
= \sup\runk \left|\frac{1}{n}\sum\runi \ke(\Zmpi) - \E\left[\frac{1}{n}\sum\runi \ke(\Zmpi)\right]\right|.
\end{align*}
Now, applying Fact~\ref{paper:fact:rademacher-bound} to these expressions, we obtain the assertions of the lemma.
\end{proof}
\subsubsection{Facts}
 Here, we state some facts used in the proof of Theorem~\ref{paper:thm:1:restatement}.
The following is Taylor's formula with the integral form of the remainder, stated using the multi-index notation.
\begin{fact}[Taylor's theorem; \citealp{ZorichMathematical2015}, Section~8.4.4]
Let \(\Omega \subset \Re^n\) be an open subset.
Let \(n \in \Na\), and let \(f : \Omega \to \Re\) be \(k\)-times continuously differentiable.
Then, for any \(x, u \in \Omega\) such that \(x + tu \in \Omega\) for all \(t \in [0, 1]\), the following equality holds:
\begin{align*}
f(x+u) - f(x) = \sum_{1 \leq |\alpha|<k} \frac{\Deriv{\alpha}{f}(x)}{\alpha!}u^\alpha + \sum_{|\alpha| = k} \frac{|\alpha|}{\alpha!} u^\alpha \int_0^1 (1 - t)^{|\alpha|-1} \Deriv{\alpha}{f}(x+tu) \dt.
\end{align*}
\label{fact:taylor}
\end{fact}
The following elementary inequality is easily proved by using the strict convexity and the strict monotonicity of the logarithm function.
\begin{fact}[Weighted AM-GM inequality]
Let \(n \in \Na\), \(x_1, \ldots, x_n \geq 0\), and \(w_1, \ldots, w_n \geq 0\).
Define \(w := w_1 + \cdots + w_n\) and assume \(w > 0\). Then,
\begin{align*}
\frac{w_1 x_1 + \cdots + w_n x_n}{w} \geq \left(x_1^{w_1} \cdots x_n^{w_n}\right)^{\frac{1}{w}}.
\end{align*}
\label{fact:AM-GM}
\end{fact}
The following standard Rademacher complexity bound is essentially due to McDiarmid's inequality, which is applied twice with the union bound~\citep[Theorem~3.3]{MohriFoundations2018}.
\begin{fact}[Rademacher complexity bound; Theorem~3.3 in \citealp{MohriFoundations2018}]
Let \(B > 0\) and \(m \in \Na\). Let \(\mathcal{G}\) be a family of functions mapping from \(\mathcal{Z}\) to \([0, B]\), and let \(z\) be a \(\mathcal{Z}\)-valued random variable.
Then, for any \(\delta > 0\), with probability at least \(1 - \delta\) over the draw of an independent and identically distributed sample \(\{z_i\}_{i=1}^m \overset{\text{i.i.d.}}{\sim} z\), the following holds:
\begin{align*}
\sup_{g \in \mathcal{G}}\left|\frac{1}{m} \sum_{i=1}^m g(z_i) - \E[g(z)]\right| \leq 2 \Rademacher{m}{\p}(\mathcal{G}) + B\sqrt{\frac{\log(2/\delta)}{2m}}.
\end{align*}
\label{paper:fact:rademacher-bound}
\end{fact}

\subsubsection{Basic Lemmas}
 Here, we prove the basic lemmas used in the proof of Theorem~\ref{paper:thm:1:restatement}.
\begin{lemma}[Convolution error bound for \HolderName{} class]
Let \(d \in \Na\), \(\beta > 1\), and \(L > 0\).
Assume that the kernel function \(K: \Re^d \to \Re\) is of order \(k = \floor{\beta}\) and satisfies
\begin{align*}
\int_{\Re^d} |K{}(u)| \cdot \|u\|^{\beta}\dr{}u < \infty.
\end{align*}
Let \(\mat{H} = \DiagMatrix{h_1, \ldots, h_d}\) with \(h_1, \ldots, h_d > 0\), and define
\(K_\mat{H}(u) := \frac{1}{|\det\mat{H}|}K(\mat{H}^{-1}u)\).
Then, for any \(f \in \HolderClass{\beta}{L}\), the following holds:
\begin{align*}
\sup_{\x \in \Re^d} \left|f(\x) - \left(\convolution{K_\mat{H}}{f}\right)(\x)\right|
\leq \HolderBoundConst{\beta}{L}{K} \opnrm{\mat{H}}^\beta,
\end{align*}
where \(\HolderBoundConst{\beta}{L}{K}\) is defined as
\begin{align*}
\HolderBoundConst{\beta}{L}{K} := L \left(\int_0^1 (1 - t)^{k-1}t^{\beta - k}\dt\right) \sum_{|\alpha| = k} \frac{\|\alpha\|^k}{\alpha! k^{k-1}} \int_{\Re^d} |K{}(u)| \cdot \|u\|^{\beta}\dr{}u
\end{align*}
and \(\alpha \in \MultiIndexSpace{d}\) runs over multi-indices.
\label{lem:Holder-conv-bound}
\end{lemma}
\begin{proof}
First, we fix \(x \in \Re^d\).
We apply the change of variables formula and obtain
\begin{align*}
&|f(x) - (\convolution{K_\mat{H}}{f})(x)| = \left|f(x) - \int_{\Re^d} K{}(u)f(x - \mat{H}u) \du\right| \tag{\text{*}}.
\end{align*}
We apply Fact~\ref{fact:taylor} to obtain
\begin{align*}
\text{(*)}&= \left|f(x) - \int_{\Re^d} K{}(u)\left(f(x) + \sum_{1 \leq |\alpha|<k} \frac{\Deriv{\alpha}{f}(x)}{\alpha!}(-\mat{H}u)^\alpha + \sum_{|\alpha| = k} \frac{|\alpha|}{\alpha!} (-\mat{H}u)^\alpha \int_0^1 (1 - t)^{|\alpha|-1} \Deriv{\alpha}{f}(x+t(-\mat{H}u)) \dt\right)\du\right| \\
&= \left|\int_{\Re^d} K{}(u)\left(\sum_{|\alpha| = k} \frac{|\alpha|}{\alpha!} (-\mat{H}u)^\alpha \int_0^1 (1 - t)^{|\alpha|-1} \Deriv{\alpha}{f}(x-t\mat{H}u) \dt\right)\du\right| \\
&= \left|\int_{\Re^d} K{}(u)\left(\sum_{|\alpha| = k} \frac{|\alpha|}{\alpha!} (-\mat{H}u)^\alpha \int_0^1 (1 - t)^{|\alpha|-1} (\Deriv{\alpha}{f}(x-t\mat{H}u) - \Deriv{\alpha}{f}(x)) \dt\right)\du\right| \\
&\leq \int_{\Re^d} |K{}(u)|\left(\sum_{|\alpha| = k} \frac{|\alpha|}{\alpha!} |\mat{H}u|^\alpha \int_0^1 (1 - t)^{|\alpha|-1} |\Deriv{\alpha}{f}(x-t\mat{H}u) - \Deriv{\alpha}{f}(x)| \dt\right)\du \tag{\text{**}},
\end{align*}
where \(\alpha = (\alpha_1, \ldots, \alpha_d)\) is a multi-index and \(|\mat{H}u|^\alpha := |h_1 u_1|^{\alpha_1} \cdots |h_d u_d|^{\alpha_d}\).
Now, by the \HolderName{}-condition of \(\Deriv{\alpha}{f}\), we have \(|\Deriv{\alpha}{f}(x-t\mat{H}u) - \Deriv{\alpha}{f}(x)| \leq L \twonrm{t\mat{H}u}^{\beta - k}\).
Also, by applying Fact~\ref{fact:AM-GM}, we have
\begin{align*}
|\mat{H}u|^\alpha &= |h_1 u_1|^{\alpha_1} \cdots |h_d u_d|^{\alpha_d}
\leq \left(\frac{1}{|\alpha|} \sum_{j=1}^d \alpha_j |h_j u_j|\right)^{|\alpha|}
\leq \left(\frac{1}{|\alpha|} \twonrm{\alpha}\cdot\twonrm{h u}\right)^{|\alpha|}
= \frac{\twonrm{\alpha}^k}{k^k} \twonrm{h u}^{k}.
\end{align*}
By applying these inequalities and imputing \(|\alpha| = k\), we obtain
\begin{align*}
\text{(**)} &\leq \int_{\Re^d} |K{}(u)|\left(\sum_{|\alpha| = k} \frac{\twonrm{\alpha}^k}{\alpha! k^{k-1}} \twonrm{\mat{H}u}^k \int_0^1 (1 - t)^{k-1} L \twonrm{t\mat{H}u}^{\beta - k} \dt\right)\du \\
&= L \left(\int_0^1 (1 - t)^{k-1}t^{\beta - k}\dt\right) \sum_{|\alpha| = k} \frac{\twonrm{\alpha}^k}{\alpha! k^{k-1}} \int_{\Re^d} |K{}(u)| \cdot \twonrm{\mat{H}u}^{\beta}\du.
\end{align*}
Finally, applying \(\twonrm{\mat{H}u} \leq \opnrm{\mat{H}}\twonrm{u}\), we have the assertion.
\end{proof}
\begin{lemma}[Bounded weights]
For all \(j \in [\D]\),
\begin{align*}
\sum\runn{i_1}\cdots\sum\runn{i_j} \wbij \cdots \wbione \in \{0, 1\}.
\end{align*}
\label{paper:lem:weights}
\end{lemma}
\begin{proof}
By direct computation, we have for any \(\setzmp\),
\begin{align*}
\sum\runi\wij(\zmp)
&= \begin{cases}
   \sum\runi \frac{1}{n} & \text{ if } \mpi{j} = \emptyset, \\
   \sum\runi 0 & \text{ if } \KHj(\zmp-\Zmpi) = 0, \forall i, \\
   \sum\runi \frac{\KHj(\zmp-\Zmpi)}{\sum\runi \KHj(\zmp-\Zmpi)} & \text{ otherwise},
   \end{cases} \\
&\in \{0, 1\}.
\end{align*}
For \(j=1\), since \(\mpi{1} = \emptyset\), we can directly show the assertion as
\begin{align*}
\sum_{i_1 = 1}^n \wbione = \sum_{i_1 = 1}^n \frac{1}{n} = 1.
\end{align*}
For \(j \geq 2\),
\begin{align*}
\sum\runn{i_1}\cdots\sum\runn{i_j} \wbij \cdots \wbione
&= \sum\runn{i_1}\cdots\sum\runn{i_{j-1}} \wbijm \cdots \wbione \left(\sum\runn{i_j} \wbij\right) \\
&\in \left\{0, \left(\sum\runn{i_1}\cdots\sum\runn{i_{j-1}} \wbijm \cdots \wbione\right)\right\}.
\end{align*}
By recursively applying the above argument for a finite number of times, we obtain the assertion for all \(j \in [\D]\).
\end{proof}

\subsection{Supplementary Theory: Comparison of Complexity Measures}
\label{sec:org0e02f7c}
\label{paper:sec:appendix:theory-2}
Here, we formally demonstrate the complexity reduction effect explained in Section~\ref{paper:sec:theoretical-insights}.
More concretely, as an example in which the effect can be demonstrated, we take the example represented by Assumption~\ref{paper:assumption:complexity} where the Lipschitz continuity of the functions are assumed
and compare the upper bounds on the complexity terms appearing in the generalization error bound of the usual empirical risk minimization (ERM) and those in Theorem~\ref{paper:thm:1} (namely \(\RadHypoK\) and \(\RadK\)).

The complexity reduction effect in this example is demonstrated by the different dependencies of the upper bounds on the sample size, both derived based on the metric-entropy method; the one corresponding to ERM yields a bound of order \(O(n^{-1/(2+D)})\) whereas the one for the proposed method yields \(O(n^{-1/3})\).
Although the comparison between the two upper bounds only provides circumstantial evidence, we believe that the reduced exponent demonstrates the complexity reduction effect as they are derived based on the same proof technique.

First, recall that the proposed method enjoys Theorem~\ref{paper:thm:1:restatement} which states,
for any \(\delta \in (0, 1)\), we have with probability at least \(1 - 2\D\delta\),
\begin{align*}
\Risk(\hf) - \Risk(\fstar) \leq 2(\ConstSumHj + \ConstSumIj) + \annot{4 \ConstMaxEpsK(\RadHypoK + \lossBound\RadK)}{Complexity terms} + 2D\lossBound\keBound\ConstMaxEpsK\sqrt{\frac{\log(2/\delta)}{2n}}.
\end{align*}

On the other hand, the usual empirical risk minimization algorithm enjoys the following theoretical guarantee.
Recall \(\hRemp(f) := \frac{1}{n} \sum\runi \ell(\f, \Zs_i)\).
\begin{proposition}[]
For any \(\delta \in (0, 1)\), with probability at least \(1 - \delta\), we have that the solution to the usual empirical risk minimization
\begin{equation*}\begin{aligned}
\hfemp \in \argmin_{\f \in \HypoCls} \{\hRemp(\f)\}
\end{aligned}\end{equation*}
satisfies
\begin{align*}
\Risk(\hfemp) - \Risk(\fstar) \leq \annot{4 \Radnp(\lF)}{Complexity term} + 2 \lossBound \sqrt{\frac{\log (2/\delta)}{2n}}.
\end{align*}
\label{prop:ERM-excess-risk-bound}
\end{proposition}
\begin{proof}
The assertion is immediate from Fact~\ref{paper:fact:rademacher-bound} and the following inequality:
\begin{equation*}\begin{aligned}
\Risk(\hfemp) - \Risk(\fstar) &= \Risk(\hfemp) - \hRemp(\hfemp) + \hRemp(\hfemp) - \Risk(\fstar) \\
&\leq \Risk(\hfemp) - \hRemp(\hfemp) + \hRemp(\fstar) - \Risk(\fstar) \leq 2 \supf |\Risk(\f) - \hRemp(\f)|.
\end{aligned}\end{equation*}
\end{proof}

From here, we compare the dependency of the complexity terms \(\Radnp(\lF)\) and \(\RadHypoK + \lossBound\RadK\) on \(n\).
In addition to Assumptions~\ref{paper:assumption:boundedness} and \ref{paper:assumption:smoothness}, assume the following:
\begin{assumption}[Complexity assumptions]
We assume the following:
\begin{itemize}
\item The functions in \(\lF\) are \(\LipConst{1}\)-Lipschitz continuous.
\item The functions \(\Kj\) are \(\LipConst{K,j}\)-Lipschitz continuous.
\item The functions \(\pkmpk{k}(\zk{k} | \cdot)\) are \(\LipConst{\p, k}\)-Lipschitz continuous for all \(\zk{k}\).
\end{itemize}
For simplicity, we also assume \(\mH = \DiagMatrix{(h, \ldots, h)}\).
\label{paper:assumption:complexity}
\end{assumption}
Under this assumption, we have the following:
\begin{proposition}[Comparison of the complexity measures]
Given Assumptions~\ref{paper:assumption:boundedness}, \ref{paper:assumption:smoothness}, and \ref{paper:assumption:complexity}, we have the following:
\begin{align*}
\Radnp(\lF) \leq \Order{n^{-\frac{1}{D + 2}}},\qquad
\RadHypoK + \lossBound\RadK \leq \Order{n^{-1/3}}.
\end{align*}
\label{prop:complexity-reduction}
\end{proposition}
\paragraph{Implications.}
Proposition~\ref{prop:complexity-reduction} shows that the complexity terms appearing in Theorem~\ref{paper:thm:1:restatement} has a better dependency on the sample size compared to those in Proposition~\ref{prop:ERM-excess-risk-bound}, demonstrating the complexity reduction effect in this example.
Note here that we do not claim that the proposed method yields a rate-optimal predictor, but instead, we provide Theorem~\ref{paper:thm:1} and this supplementary analysis to obtain insights regarding how the proposed method may facilitate the learning.

\begin{proof}[Proof of Proposition~\ref{prop:complexity-reduction}]
By the Lipschitz continuity of the functions in \(\lF\) and the boundedness of \(\ZspAll\), we can apply Fact~\ref{fact:Lipschitz-functions} to obtain
\begin{align*}
\log \CoveringNumber{\lF}{t}{\|\cdot\|_\infty} \leq C \left(\frac{\LipConst{1}}{t}\right)^{\D}
\end{align*}
for a constant \(C > 0\).
By applying Fact~\ref{fact:discretization-bound}, and minimizing the right-hand side for \(t\), we have the first assertion.

On the other hand, by Lemma~\ref{lem:product-entropy},
\begin{align*}
\log \CoveringNumber{\lFjOne \otimes \KernelShiftedClassHj}{t}{\|\cdot\|_\infty}
\leq \log \CoveringNumber{\lFjOne}{t_1}{\|\cdot\|_\infty} + \log \CoveringNumber{\KernelShiftedClassHj}{t_2}{\|\cdot\|_\infty},
\end{align*}
where \(t_1, t_2\) are such that \(\keBound t_1 + \lossBound t_2 = t\).
Now, applying Lemma~\ref{lem:Lipschitz-Curried},
\begin{align*}
\log \CoveringNumber{\lFjOne}{t_1}{\|\cdot\|_\infty}
\leq \log \sup_{z \in \Zspk{j-1}} \CoveringNumber{\mathcal{F}_z}{t_{1,1}}{\|\cdot\|_\infty}
+ \log\CoveringNumber{\EuclideanBall{j-1}{\ZspRadiusBound}}{t_{1,2}}{\|\cdot\|}
\end{align*}
By combining Lemma~\ref{lem:Lipschitz-marginalized} and Lemma~\ref{lem:metric-entropy-continuity-restriction},
and applying Fact~\ref{fact:euclidean-norm-entropy}, we have
\begin{align*}
\log \sup_{z \in \Zspk{j-1}} \CoveringNumber{\mathcal{F}_z}{t_{1,1}}{\|\cdot\|_\infty}
\leq C \frac{\LipConst{2}}{t_{1,1}},
\quad
\log\CoveringNumber{\EuclideanBall{j-1}{\ZspRadiusBound}}{t_{1,2}}{\|\cdot\|}
\leq (j-1) \log \left(1 + \frac{2\ZspRadiusBound}{t_{1,2}}\right),
\end{align*}
where \(t_{1,1}, t_{1,2}\) are such that \(t_1 = t_{1,1} + \LipConst{2}t_{1,2}\),
and \(\LipConst{2} = \LipConst{1} + \lossBound \sum_k \LipConst{\p, k}\).

On the other hand, by Lemma~\ref{lem:shifter-kernel-complexity},
we have
\begin{align*}
\log \CoveringNumber{\KernelShiftedClassHj}{t_2}{\|\cdot\|_\infty}
\leq \mpjsize \log \left(1 + \frac{2\LipConst{K,H,j}\ZspRadiusBound}{t_2}\right).
\end{align*}
where \(\LipConst{K,H,j} = h^{{-\mpjsize} - 1} \LipConst{K,j}\).

Therefore, we have
\begin{align*}
\log \CoveringNumber{\lFjOne \otimes \KernelShiftedClassHj}{t}{\|\cdot\|_\infty}
\leq C \frac{\LipConst{2}}{t_{1,1}}
+ (j-1) \log \left(1 + \frac{2\ZspRadiusBound}{t_{1,2}}\right)
+ \mpjsize \log \left(1 + \frac{2\LipConst{K,H,j}\ZspRadiusBound}{t_2}\right).
\end{align*}

By applying Fact~\ref{fact:discretization-bound}, letting
\begin{align*}
t_{1,1} = \frac{t}{3 \keBound}, \ t_{1,2} = \frac{t}{3 \keBound \LipConst{2}}, \ t_2 = \frac{t}{3 \lossBound},
\end{align*}
and minimizing the upper bound for \(t\), we have
\begin{align*}
\absdet{\Hj}\lFjOneKRad \leq \Order{n^{-1/3}}.
\end{align*}
Therefore, we have
\begin{align*}
\RadHypoK &= \sum\runj \absdet{\Hj}\lFjOneKRad \leq \Order{n^{-1/3}}, \\
\RadK &= \sum\runj \absdet{\Hj}\KRad \leq \Order{n^{-1/2}},
\end{align*}
and obtain the second assertion.
\end{proof}
\subsubsection{Lemmas and Facts}
\begin{lemma}[Metric entropy of products]
Let \(\mathcal{F}, \mathcal{G}\) be two classes of bounded measurable functions satisfying
\(\|f\|_\infty \leq M_\mathcal{F} (f \in \mathcal{F})\) and \(\|g\|_\infty \leq M_\mathcal{G} (g \in \mathcal{G})\).
Then, we have for any \(t_1, t_2 > 0\),
\begin{align*}
\log \CoveringNumber{\mathcal{F} \otimes \mathcal{G}}{t}{\|\cdot\|_\infty}
\leq \log \CoveringNumber{\mathcal{F}}{t_1}{\|\cdot\|_\infty} + \log \CoveringNumber{\mathcal{G}}{t_2}{\|\cdot\|_\infty}
\end{align*}
where \(t = M_\mathcal{G} t_1 + M_\mathcal{F} t_2\).
\label{lem:product-entropy}
\end{lemma}
\begin{proof}
Let \(\{f_i\}_i\) (\(\{g_j\}_j\)) be the \(t_1\)- (resp. \(t_2\)-)covering of \(\mathcal{F}\) (resp. \(\mathcal{G}\)).
Then, for any \(f \in \mathcal{F}\) and \(g \in \mathcal{G}\), we have for some \(i, j\) that
\begin{align*}
\|f \otimes g - f_i \otimes g_j\|_\infty
&\leq \|f \otimes g - f_i \otimes g\|_\infty + \|f_i \otimes g - f_i \otimes g_j\|_\infty \\
&\leq \|f - f_i\|_\infty M_\mathcal{G} + M_\mathcal{F} \|g - g_j\|_\infty \\
&\leq M_\mathcal{G} t_1 + M_\mathcal{F} t_2.
\end{align*}
This implies the assertion.
\end{proof}
\begin{lemma}[Lipschitz continuity of marginalized function class]
Assume that \(\pkmpk{k}(\zk{k} | \cdot)\) is \(\LipConst{\p, k}\)-Lipschitz continuous for all \(\zk{k}\).
Then, the elements of \(\lFj\) are Lipschitz continuous with the constant \(\LipConst{1} + \lossBound \sum_k \LipConst{\p, k}\).
\label{lem:Lipschitz-marginalized}
\end{lemma}
\begin{proof}
Since the functions in \(\lF\) are \(\LipConst{1}\)-Lipschitz continuous,
the elements of \(\lFj\) are also Lipschitz continuous:
\begin{equation*}\begin{aligned}
|\lfj(x) - \lfj(y)| &= \left|\int \lf((x, z)) \prod_k \pkmpk{k}(\zk{k}|(x, z)^{\mpi{k}})\dz - \int \lf((y, z))\prod_k \pkmpk{k}(\zk{k}|(y, z)^{\mpi{k}})\dz\right| \\
&\leq \int |\lf((x, z)) - \lf((y, z))| \prod_k \pkmpk{k}(\zk{k}|(y, z)) \dz \\
&\quad+ \sum_{k \geq j+1} \int |\lf((x, z))| \pkmpk{j+1}(\zk{j+1}|(x, z)) \cdots (\pkmpk{k}(\zk{k} | (x, z)) - \pkmpk{k}(\zk|(y, z))) \cdots \pkmpk{D}(\zk{D}|(y, z)) \dz \\
&\leq \LipConst{1} \|x - y\| \cdot 1 + \lossBound \sum_k 1 \cdot \LipConst{\p, k} \|x - y\| \cdot 1 \\
&\leq (\LipConst{1} + \lossBound \sum_k \LipConst{\p, k}) \|x - y\|.
\end{aligned}\end{equation*}
\end{proof}
\begin{lemma}[Lipschitz continuity of curried function class]
Let \(j \in [2:\D]\) and \(\ZspRadiusBound = \sup_{z \in \Zsp}\|z\|\).
Also let \(\EuclideanBall{j-1}{R}\) denote the radius-\(R\) ball in the \((j-1)\)-dimensional Euclidean space,
and define \(\mathcal{F}_z := \{\lfj(z, \cdot) : \lfj \in \lFj\}\) for \(z \in \Zspk{j-1}\).
Assume \(\lFj\) consist of \(\LipConst{2}\)-Lipschitz continuous functions.
Then, we have
\begin{align*}
\log \CoveringNumber{\lFjOne}{t}{\|\cdot\|_\infty}
\leq \log \sup_{z \in \Zspk{j-1}} \CoveringNumber{\mathcal{F}_z}{u}{\|\cdot\|_\infty}
+ \log\CoveringNumber{\EuclideanBall{j-1}{\ZspRadiusBound}}{v}{\|\cdot\|}
\end{align*}
where \(t, u, v > 0\) are such that \(t = u + \LipConst{2} v\).
\label{lem:Lipschitz-Curried}
\end{lemma}
\begin{proof}
Let \(\{z_\mu\}_{\mu} \subset \Zspk{j-1}\) be a \(v\)-covering of \(\Zspk{j-1}\).
For each \(z_\mu\), consider the set \(\mathcal{F}_{\mu} = \{\lfj(z_\mu, \cdot) : \lfj \in \lFj\}\).
Let \(\{\lfj^{\mu, k}\}_k \subset \mathcal{F}_{\mu}\) be a \(u\)-covering of \(\mathcal{F}_{\mu}\).
Then, for any \(\lfj \in \lFj\) and \(z \in \Zspk{j-1}\),
there exists \(z_\mu\) such that \(\|z_\mu - z\| \leq v\).
Moreover, since we have \(\lfj(z_\mu, \cdot) \in \mathcal{F}_{\mu}\),
there exists \(\lfj^{\mu, k}\) such that \(\|\lfj(z_\mu, \cdot) - \lfj^{\mu, k}(z_\mu, \cdot)\|_\infty \leq u\).
For such a pair \((z_\mu, \lfj^{\mu, k})\), we have
\begin{align*}
\|\lfj(z, \cdot) - \lfj^{\mu, k}(z_\mu, \cdot)\|_\infty
&\leq \|\lfj(z, \cdot) - \lfj(z_\mu, \cdot)\|_\infty + \|\lfj(z_\mu, \cdot) - \lfj^{\mu, k}(z_\mu, \cdot)\|_\infty
\leq \LipConst{2} v + u
\end{align*}
Therefore, the set \(\bigcup_{\mu} \{z_\mu\}_{\mu} \times \{\lfj^{\mu, k}\}_k\) induces a \((\LipConst{2} v + u)\)-covering of \(\lFjOne\).
Noting that the cardinality of \(\bigcup_{\mu} \{z_\mu\}_{\mu}\) is bounded by \(\CoveringNumber{\EuclideanBall{j-1}{\ZspRadiusBound}}{v}{\|\cdot\|}\)
and that of \(\{\lfj^{\mu, k}\}_k\) by \(\sup_{z \in \Zspk{j-1}} \CoveringNumber{\mathcal{F}_z}{u}{\|\cdot\|_\infty}\),
we have the assertion.
\end{proof}
\begin{lemma}[Metric entropy of functions curried by a specific input]
Assume that the elements of \(\lFj\) are \(\LipConst{2}\)-Lipschitz continuous.
Then, there exists a constant \(C > 0\) such that for sufficiently small \(u > 0\),
\begin{align*}
\sup_{z \in \Zspk{j-1}} \CoveringNumber{\mathcal{F}_z}{u}{\|\cdot\|_\infty}
\leq C \frac{\LipConst{2}}{u}.
\end{align*}
\label{lem:metric-entropy-continuity-restriction}
\end{lemma}
\begin{proof}
Since the elements of \(\lFj\) are \(\LipConst{2}\)-Lipschitz continuous, so are the elements of \(\mathcal{F}_{z}\) with Lipschitz constant \(\LipConst{2}\).
Indeed, for any \(x, y \in \Zspk{j}\) and \(z \in \Zspk{j-1}\), we have
\begin{align*}
|\lfj(z, x) - \lfj(z, y)| \leq \LipConst{2}\left\|\begin{pmatrix} z \\ x \end{pmatrix} - \begin{pmatrix} z \\ y \end{pmatrix}\right\|
= \LipConst{2} \|x - y\|.
\end{align*}
Therefore, by applying Lemma~\ref{fact:Lipschitz-functions}, we have the assertion.
\end{proof}
\begin{lemma}[Shifted kernel complexity]
Assume that \(\Kj : \Rempj \to \Re\) is \(\LipConst{K,j}\)-Lipschitz continuous.
Let \(\LipConst{K,H,j} = \frac{1}{|\det\Hj|} \LipConst{K,j} \opnrm{\Hj^{-1}}\).
Then, we have the following:
\begin{align*}
\log \CoveringNumber{\KernelShiftedClassHj}{t_2}{\|\cdot\|_\infty}
\leq \mpjsize \log \left(1 + \frac{2\LipConst{K,H,j}\ZspRadiusBound}{t_2}\right).
\end{align*}
\label{lem:shifter-kernel-complexity}
\end{lemma}
\begin{proof}
Recalling \(\KHj(u) = \frac{1}{|\det\Hj|}\Kj(\Hj^{-1} u)\),
for any \(\KHj(z_1 - \cdot), \KHj(z_2 - \cdot) \in \KernelShiftedClassHj\), we have
\begin{align*}
\|\KHj(z_1 - \cdot) - \KHj(z_2 - \cdot)\|_\infty
\leq \frac{1}{|\det\Hj|} \LipConst{K,j} \|\Hj^{-1}(z_1 - z_2)\| \\
\leq \frac{1}{|\det\Hj|} \LipConst{K,j} \opnrm{\Hj^{-1}} \|z_1 - z_2\|
\end{align*}
Therefore, we have
\begin{align*}
\log \CoveringNumber{\KernelShiftedClassHj}{t_2}{\|\cdot\|_\infty}
\leq \log \CoveringNumber{\Zmpj}{t_2/\LipConst{K,H,j}}{\|\cdot\|}.
\end{align*}
Applying Fact~\ref{fact:euclidean-norm-entropy}, we obtain the assertion.
\end{proof}
\begin{fact}[One-step discretization bound]
Let \(\mathcal{F}\) be a class of measurable functions.
There exist constants \(c\) and \(B\) such that for any \(t \in (0, B]\), the following relation between the Rademacher complexity and the metric entropy holds:
\begin{align*}
\Radmq(\mathcal{F}) \leq t + c \sqrt{\frac{\log \CoveringNumber{\mathcal{F}}{t}{\|\cdot\|_\infty}}{m}}
\end{align*}
\label{fact:discretization-bound}
\end{fact}

\begin{fact}[Euclidean ball metric entropy bound; \citealp{WainwrightHighDimensional2019}, Example~5.8, p.126]
Let \(R > 0\) and \(d \in \Na\).
Let \(\mathcal{B}(R)\) denote the radius-\(R\) ball in the \(d\)-dimensional Euclidean space.
Then, we have the following metric entropy bound:
\begin{align*}
\log \CoveringNumber{\mathcal{B}(R)}{\delta}{\|\cdot\|} \leq d \log\left(1 + \frac{2R}{\delta}\right).
\end{align*}
\label{fact:euclidean-norm-entropy}
\end{fact}

\begin{fact}[Lipschitz functions metric entropy bound; \citealp{WainwrightHighDimensional2019}, Example~5.10, p.129]
Let \(L, R > 0\) and \(d \in \Na\).
Let \(\mathrm{Lip}(R, L)\) denote the set of \(L\)-Lipschitz functions on \([0, R]^d\).
Then, we have the following metric entropy bound for sufficiently small \(\delta > 0\):
\begin{align*}
\log \CoveringNumber{\mathrm{Lip}(R, L)}{\delta}{\|\cdot\|_\infty}
\leq C \left(\frac{LR}{\delta}\right)^d,
\end{align*}
where \(C > 0\) is a constant.
\label{fact:Lipschitz-functions}
\end{fact}
\section{Computational complexity of Algorithm~\ref{paper:alg:proposed-method}}
\label{sec:org224d4a3}
\label{paper:sec:appendix:computational-complexity}
Here, we remark why the worst-case computational complexity of Algorithm~\ref{paper:alg:proposed-method} is \(\CompOrder{n^\D}\).
The main computation cost of Algorithm~\ref{paper:alg:proposed-method} comes from the computation of the weights \(\wbij\).
There are \(n^{j-1}\) nodes at depth \(j\)
(\Figure{\ref{fig:probability-tree}}),
each with \(n\) weighted edges connected to depth \(j+1\).
The set of weights corresponding to each node, \(\{\wbij\}_{i_j \in [n]}\),
is computed by constructing a matrix of shape \(n \times n^{j-1}\) each of whose element is the kernel value for two vectors of dimensionality \(\mpjsize (\leq j - 1)\).
In the case of Gaussian kernels, each kernel value requires \(\CompOrder{j - 1}\) operations to compute.
Subsequently, the kernel matrix is normalized by the column sum, which requires \(\CompOrder{n}\) summations and \(n^j\) divisions.
The same computation takes place for each of the \(\bijm \in [n]^{j-1}\) nodes at depth \(j\),
therefore, the edge weights between depth \(j\) and depth \(j+1\) can be computed by \(\CompOrder{n^j}\) operations.
The edge weights are multiplied to obtain the node weights, which requires \(\CompOrder{n^\D}\) multiplications since the number of multiplications that take place is equal to the number of edges in \Figure{\ref{fig:probability-tree}}.
Overall, Algorithm~\ref{paper:alg:proposed-method} requires \(\CompOrder{n^\D}\) operations for the edge weight computation and \(\CompOrder{n^\D}\) for the node weight computation, amounting to \(\CompOrder{n^\D}\) operations in total, in the worst case that no edge is pruned by the threshold \(\theta\).

\end{appendices}
\end{document}